\documentclass[10pt,twocolumn,letterpaper]{article}

\usepackage[accsupp]{axessibility} % Improves PDF readability for those with visual impairments.
\usepackage{marvosym}
\usepackage{cvpr}              % To produce the CAMERA-READY version
\usepackage[dvipsnames]{xcolor}

\definecolor{cvprblue}{rgb}{0.21,0.49,0.74}
\usepackage[pagebackref,breaklinks,colorlinks,citecolor=cvprblue]{hyperref}

\def\paperID{15924} % *** Enter the Paper ID here
\def\confName{CVPR}
\def\confYear{2024}

\title{Leveraging Cross-Modal Neighbor Representation \\ for Improved CLIP Classification}

\author{Chao Yi, Lu Ren, De-Chuan Zhan, Han-Jia Ye\textsuperscript{(\Letter)}\\
National Key Laboratory for Novel Software Technology, Nanjing University, China\\
School of Artificial Intelligence, Nanjing University, China\\
{\tt\small \{yic,renl,zhandc,yehj\}@lamda.nju.edu.cn}
}

\usepackage{amssymb}
\usepackage{amsmath,amsfonts}
\usepackage{amsopn}
\usepackage{bm} % bold symbol
\usepackage{multirow}
\usepackage{flushend}
\usepackage{tabularx}
\newcommand{\vct}[1]{\boldsymbol{#1}} % vector
\newcommand{\mat}[1]{\boldsymbol{#1}} % matrix
\newcommand{\ProbOpr}[1]{\mathbb{#1}}
\newcommand{\expect}[2]{%
\ifthenelse{\equal{#2}{}}{\ProbOpr{E}_{#1}}
{\ifthenelse{\equal{#1}{}}{\ProbOpr{E}\left[#2\right]}{\ProbOpr{E}_{#1}\left[#2\right]}}} % Expectation: syntax: E{1}{2} = E_1[2], E{}{2}=E[2], E{1}{} = E_1
\newcommand{\x}{{\vct{x}}}

\newcommand{\vt}{\vct{t}}

\newcommand{\mW}{\mat{W}}

\newcommand{\eat}[1]{}

\newcommand{\bbR}{\mathbb{R}}

\usepackage{xspace}
\usepackage{xcolor}
\usepackage{csquotes}
\usepackage{adjustbox}
\usepackage{float}
\usepackage{longtable}
\usepackage{array}
\newcommand{\ATG}{{\sc ATG}\xspace}

\newcommand{\CODER}{{\sc CODER}\xspace}

\begin{document}
\maketitle

\begin{abstract}
CLIP showcases exceptional cross-modal matching capabilities due to its training on image-text contrastive learning tasks. However, without specific optimization for unimodal scenarios, its performance in single-modality feature extraction might be suboptimal. Despite this, some studies have directly used CLIP's image encoder for tasks like few-shot classification, introducing a misalignment between its pre-training objectives and feature extraction methods. This inconsistency can diminish the quality of the image's feature representation, adversely affecting CLIP's effectiveness in target tasks. In this paper, we view text features as precise neighbors of image features in CLIP's space and present a novel \textbf{C}r\textbf{O}ss-mo\textbf{D}al n\textbf{E}ighbor \textbf{R}epresentation~(\CODER) based on the distance structure between images and their neighbor texts. This feature extraction method aligns better with CLIP's pre-training objectives, thereby fully leveraging CLIP's robust cross-modal capabilities. The key to construct a high-quality \CODER lies in how to create a vast amount of high-quality and diverse texts to match with images. We introduce the \textbf{A}uto \textbf{T}ext \textbf{G}enerator~(\ATG) to automatically generate the required texts in a data-free and training-free manner. We apply \CODER to CLIP's zero-shot and few-shot image classification tasks. Experiment results across various datasets and models confirm \CODER's effectiveness. 
Code is available at: \url{https://github.com/YCaigogogo/CVPR24-CODER}.
\end{abstract}

\section{Introduction}
In recent years, vision-language models have garnered widespread attention, with CLIP~\cite{CLIP} standing out as a notably powerful exemplar. Trained on a vast array of image-text pairs through the image-text contrastive learning tasks, CLIP boasts impressive cross-modal matching capabilities. And it has been applied to fields like image classification~\cite{CLIP}, object detection~\cite{obj1}, semantic segmentation~\cite{seg1,seg2}, video understanding~\cite{clip4clip}, voice classification~\cite{audioclip}, text-to-image generation~\cite{t2i1, t2i2}, model pretraining~\cite{mvp}, and beyond~\cite{clip-vil,pointclip}.

Some existing works~\cite{tip-adapter,mvp,clipasso} extract image features directly from CLIP's image encoder for intra-modal operation, like calculating similarity between images in few-shot classification~\cite{tip-adapter}. However, these methods overlook CLIP's multi-modal capabilities, leading to a misalignment with CLIP's pre-training objectives. Furthermore, since CLIP isn't optimized for uni-modal scenarios, its performance in intra-modal tasks isn't guaranteed. To optimize the image features extracted by CLIP, we ask:
\begin{displayquote}
Can we leverage CLIP's powerful cross-modal matching capabilities to extract better image features, thereby improving CLIP's performance on downstream tasks?
\end{displayquote}

\begin{figure*}
  \centering
  \includegraphics[width=0.9\linewidth]{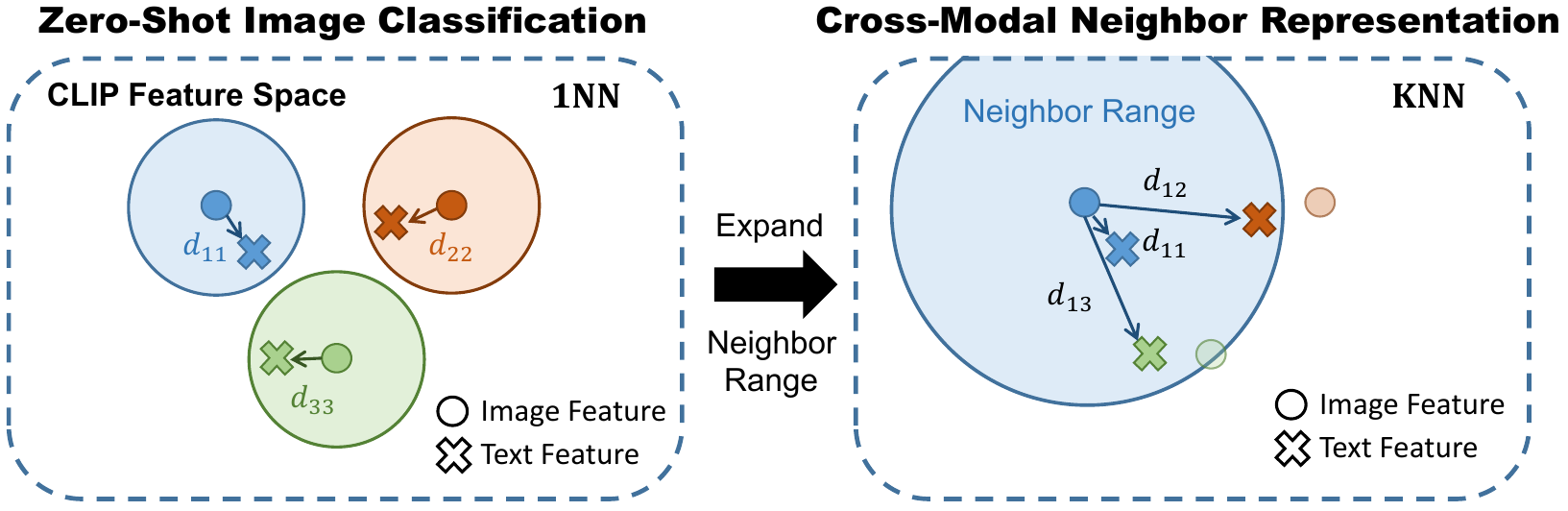} 
  \caption{\textbf{Illustration of image's \textbf{C}r\textbf{O}ss-mo\textbf{D}al n\textbf{E}ighbor \textbf{R}epresentation~({\scshape Coder})}. CLIP's powerful text-image matching capabilities endows it with a favorable cross-modal neighbor distance relation. And CLIP's Zero-Shot Image Classification process can be interpreted as using a $1$NN algorithm to find the image's nearest text, with the text's class determining the image's predicted class. Inspired by this idea, we expand the image's neighbor range to leverage its distance to all texts for constructing the \CODER. Here $d_{ij}$ refers to the distance between the $i$-th image and the $j$-th text.}
  \label{fig.1}
  \vspace{-0.3cm}
\end{figure*}

In this paper, we introduce an enhanced image representation based on the distance between images and their neighboring texts in CLIP's feature space. This idea stems from our re-examination of CLIP's robust zero-shot classification capability from the perspective of nearest neighboring: The previous approaches consider the text features extracted by CLIP as classifiers and use them to get the classification results. Different from this perspective, we interpret CLIP's zero-shot image classification as a $1$NN problem, as shown in the left part of \autoref{fig.1}. We treat texts as the images' neighbors in the CLIP's feature space. Then for each image, CLIP identifies its closest text and assigns this text's class as the image's predicted class. This $1$NN approach shows good performance because CLIP's robust image-text matching capabilities ensure images are closer to their semantically related texts. This suggests that the cross-modal distance between an image and its neighboring texts captures information about the image, such as its class.

In zero-shot image classification, CLIP only considers the distance between an image and its nearest neighbor text, which loses the information implied in the distance between the image and other texts. To make full use of this information, we expand each image's neighbor range to utilize its distance to $\bm{K}$ \textbf{N}earest \textbf{N}eighbor~($K$NN) texts for constructing image representation, as depicted in the right half of \autoref{fig.1}. Here $K$ denotes the total number of texts. We refer to this representation as \textbf{C}r\textbf{O}ss-mo\textbf{D}al n\textbf{E}ighbor \textbf{R}epresentation~(\CODER). We believe images with closer \CODER values are more similar in semantics. This aligns with intuition: If two objects share the same sets of similar and dissimilar items, they're likely similar to each other.

Previous work~\cite{wjx1,wjx2} has noted that dense sampling of neighbor samples is critical for building neighbor representations. This inspires us to use various high-quality texts for dense sampling. Considering that Large Models have rich knowledge and are widely used in many ways~\cite{llm1,llm2,llm3}, we introduce the \textbf{A}uto \textbf{T}ext \textbf{G}enerator~(\ATG) based on Large Language Models like ChatGPT~\cite{InstructGPT} to automatically generate various high-quality texts.
Without the need for image data and the training process, \ATG can produce a diverse and effective set of texts based on the target dataset's class names. These diverse, high-quality texts enhance the density of neighboring texts for images in CLIP's feature space, helping to build a better \CODER.

We apply \CODER to CLIP's zero-shot and few-shot image classification tasks to prove \CODER's superiority. For the zero-shot image classification task, we use \CODER in a two-stage manner. In the first stage, we use \ATG to acquire general class-specific texts for constructing general \CODER. Then we employ a heuristic classifier to the image's general \CODER. In the second stage, we utilize \ATG to generate one-to-one specific texts for comparing two distinct classes, concentrating on their most distinguishing features. These texts are then used to construct the one-to-one specific \CODER, which is used to rerank the preliminary classification results. For the few-shot situation, we calculate the similarity between test images and the support set's images using those images' general \CODER. By ensembling the similarity and CLIP's original zero-shot classification logits, we determine the classification results. Experiment results on various datasets and different CLIP models confirm that \CODER enhances CLIP's performance in both zero-shot and few-shot image classification.

\section{Related Work}
\noindent\textbf{Vision-Language Models.} Vision-Language Models (VLMs) represent a class of multimodal models adept at correlating textual and visual information.
Prominent models in this domain include CLIP~\citep{CLIP}, ALIGN~\citep{align}, FLAVA~\citep{flava} and Florence~\citep{florence}, among others. 
These models typically comprise two main components: an image encoder and a text encoder.
Some models also have multi-modal fusion modules~\citep{albef, beit-3}.
VLMs are trained on extensive text-image pairs through tasks like image-text contrastive learning, endowing them with powerful text-image matching capabilities. 
In this paper, we leverage these capabilities of CLIP to generate our cross-modal neighbor representations~(\CODER) for images.

\noindent\textbf{Using LLMs as Experts to Improve VLMs.} Large Language Models~(LLMs) are widely used in many tasks, such as in-context learning. LLMs can serve as experts by outputting their knowledge in text form. This text-based knowledge can be harnessed by VLMs to enhance their capabilities. 
Previous work has explored using LLMs' knowledge to optimize VLMs' pre-training~\citep{rewrite_clip}, prediction interpretability~\citep{vcd,cbm1,cbm2}, and classifier quality~\citep{vcd,bottom-up,cupl,double_dought}.
Recently, some work~\cite{LOVM,select_vlm} uses LLMs to select the best VLMs from a VLM zoo for specific target tasks.
In this paper, we use the semantic information provided by LLMs to optimize CLIP's features during the inference stage in a training-free manner. 

\begin{figure*}[t]
  \centering
  \includegraphics[width=0.89\linewidth]{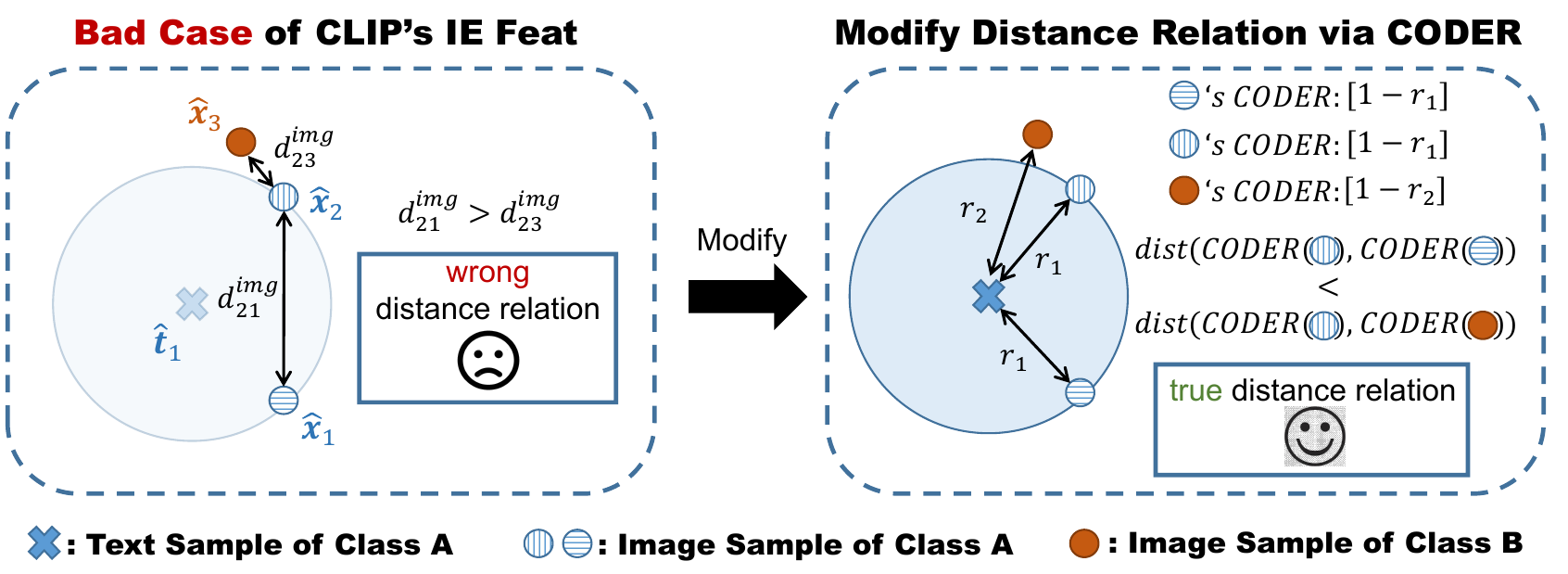} % 替换为您的图片路径
  \caption{\textbf{An example of \CODER correcting wrong distance relation between images.} $d_{ij}^{img}$ refers to the cosine distance between the $i$-th and the $j$-th image. $r_1$ and $r_2$ refer to the cosine distance between the text and different images, while $1-r_1$ and $1-r_2$ refer to the cosine similarity. The left side of the figure indicates that even though images of the same class share similar distance to a text, this doesn't ensure that their features are closely similar. The right side of the figure shows that \CODER corrects the wrong distance relation by utilizing the text-image distance.}
  \label{fig.2}
  \vspace{-0.3cm}
\end{figure*}

\section{Notations and Background}
\textbf{Using CLIP to match text and image.}
CLIP encodes both images and texts into a joint space using its image encoder $f^\textit{I}$ or text encoder $f^\textit{T}$, shown in \autoref{clip_map1}. Here $\hat{\x}_{i}$ and $\hat{\vt}_{j}$ refers to the feature of image $\x_i$ and text $\vt_j$, respectively. $D_1$ refers to the dimension of CLIP's feature space.
\begin{equation}
\label{clip_map1}
\hat{\x}_i = f^{\textit{I}}(\x_i)\in \mathbb{R}^{D_1}, \ \hat{\vt}_{j} = f^{\textit{T}}(\vt_{j})\in \mathbb{R}^{D_1}.
\end{equation}
\begin{equation}
\label{clip_match}
    \Tilde{\vt} = \mathop{\arg\max}_{j\in[1,\cdots,K]} \frac{\hat{\x}_{i}^{\top}\hat{\vt}_j}{\|\hat{\x}_{i}\| \cdot \|\hat{\vt}_{j}\|}.
\end{equation}
Then we can compute the cosine similarity between the features of various texts and images. These similarity scores represent the matching degree between different images and texts. 
Given an image $\x_i$, its best matched text $\Tilde{t}$ is identified by the highest similarity scores with the image, as shown in \autoref{clip_match}.
Here $K$ refers to the number of texts.
\section{Cross-Modal Neighbor Representation}
\subsection{Understand the Advantage of {\scshape{Coder}}.}
We construct the \CODER for the current image by tapping into the precise image-text distance relationship within the CLIP feature space. 
This process is shown in \autoref{construct}. We use the \CODER construct function $\phi$ to build image's \CODER $\phi \left(\hat{\x_i}\right)$ based on its original feature $\hat{\x_i}$.
\begin{equation}
\label{construct}
\phi \left(\hat{\x}_i\right) = \left[\psi\left(d\left(\hat{\x}_i, \hat{\vt}_1\right)\right), \cdots, \psi\left(d\left(\hat{\x}_i, \hat{\vt}_K\right)\right) \right] \in \bbR^{K}.
\end{equation}
The specific implementation of $\phi$ depends on the similarity or distance function $d$ and the subsequent mapping function $\psi$ applied to this distance. 
In this paper, we use cosine similarity for $d$ and identity mapping for $\psi$. 
Then we can rewrite \autoref{construct} as \autoref{construct_CODER}. We emphasize that the implementation of $d$ and $\psi$ can be further researched. 
\begin{equation}
\label{construct_CODER}
    \phi \left(\hat{\x}_i\right) = \left[\frac{\hat{\x}_i^{\top} \hat{\vt}_1}{\|\hat{\x}_i\| \cdot \|\hat{\vt}_1\|}, \cdots, \frac{\hat{\x}_i^{\top} \hat{\vt}_K}{\|\hat{\x}_i\| \cdot |\hat{\vt}_K\|}\right] \in \bbR^{K}.
\end{equation}
We highlight \CODER's advantages over CLIP's original image features using an example. 
\autoref{fig.2}'s left side depicts a bad case for CLIP. 
For simplicity, we consider a situation where test images share the same neighboring text in CLIP's feature space. While CLIP's cross-modal pre-training ensures accurate image-text distances, it doesn't always capture precise distances between images. This results in cases where the distance for same-class images $d_{21}^{img}$ exceeds that of different-class images $d_{23}^{img}$. To solve this problem, \CODER uses CLIP's accurate text-image distance to build image features. As images of the same class have similar distance to their neighboring texts, their \CODER align more closely. Thus, \CODER addresses the wrong distance relation between images.

We then focus on the key element of building a good \CODER. Previous studies have emphasized that dense sampling of neighboring samples is vital for algorithms based on nearest neighbor. For example, only when the training samples are densely sampled will the error rate of the $K$NN classifier remain within twice that of the Bayes optimal classifier. And some studies~\citep{wjx1,wjx2} have observed that greater sampling density of neighboring samples can lead to better neighbor representations. Inspired by these works, we try to optimize our \CODER by increasing the number of texts. For increasing the number of texts, we use our \textbf{A}uto \textbf{T}ext \textbf{G}enerator~(\ATG) to achieve this objective.

\subsection{Use Auto Text Generator to Generate Texts}
Accurate and diverse class-specific texts provide a comprehensive description of object classes from various perspectives, enhancing the sampling density of images' neighbor texts for constructing better \CODER. 
To automatically generate a plethora of high-quality class-specific texts adaptive to target tasks, we introduce the \textbf{A}uto \textbf{T}ext \textbf{G}enerator~(\ATG). 
It uses different query prompts to extract diverse insights from external experts like ChatGPT~\citep{InstructGPT}. 
Then it leverages that knowledge to construct various high-quality texts. 
These texts are instrumental in generating high-quality \CODER.

In our implementation, \ATG can construct five types of texts: (1) Class Name-based Texts; (2) Attribute-based Texts; (3) Analogous Class-based Texts; (4) Synonym-based Texts; (5) One-to-One Texts. The first two types of texts are proposed by previous work\citep{CLIP,vcd}, while the latter three are innovations introduced in this paper. We will then delve deeper into the design rationale and generation process of these last three types of texts.

(1) \textbf{Analogous Class-based Texts}. If someone describes a clouded leopard as resembling a cheetah, you can envision its appearance even if you've never seen a clouded leopard. This scenario illustrates how inter-object similarities help humans leverage their experience with known object classes to recognize new ones. Inspired by this, we query ChatGPT to obtain analogous categories for a given object by asking:
\vspace{-0.6cm}
\begin{center}
\begin{tabular}{p{8cm}}
  \small
  \texttt{\ Q: What other categories are \{class\} visually similar to?} \\
  %\texttt{\ A: \_} \\
  \end{tabular}
\end{center}

Given that the generated class names might exist within the target dataset's class names, we address this issue by filtering out those generated class names whose cosine similarity with existing class names surpass a defined threshold. Next, we insert each analogous class name into some templates like ``\texttt{a \{target class\} similar to \{analogous class\}}'' to form a complete text. 

(2) \textbf{Synonym-based Texts}. An object can have several names. For example, both ``forest'' and ``woodland'' refer to ``land covered with trees and shrubs''. Due to varying frequencies of synonyms in CLIP's training data, the model might favor more common terms and undervalue lesser-known synonyms, despite their equivalent meanings. To mitigate this bias, we query WordNet for synonyms of the current class. Subsequently, we insert the obtained synonym class names into templates like ``\texttt{a photo of \{synonym class\}}''. 

(3) \textbf{One-to-One Texts}: Due to similar classes often share common features, the \ATG may generate nearly identical attribute descriptions for these classes. For example, the \ATG might generate attributes such as ``two pairs of wings" for both ``butterfly'' and ``dragonfly''. Such identical attributes can hinder CLIP's ability to distinguish closely related classes. To tackle this problem, we introduce the one-to-one texts. We use the following query prompt to guide ChatGPT in generating the most distinguishing features between similar classes A and B:
\vspace{-0.6cm}
\begin{center}
\begin{tabular}{p{8cm}}
 \small
  \texttt{\ Q: What are different visual features between a \{class 1\} and a \{class 2\} in a photo? Focus on their key differences.} \\
  \end{tabular}
\end{center}
 Using the prompt, we generate distinguishing attributes that differentiate butterfly from dragonfly. For butterfly, some of the exemplified attributes include: [``Butterflies typically have larger and more colorful wings compared to dragonflies.'']. For dragonfly, some of the descriptors produced are: [``They have transparent wings that are typically held out horizontally when at rest.'']. From these newly created attributes, we can make two main observations:~(1)~The attributes underscore the key differences between butterfly and dragonfly, such as their wing characteristics. (2)~The new attributes accentuate the comparison between the two classes, evident from terms like ``larger'', and ``compared to'' found within the descriptors. 

 Finally, we insert the obtained one-to-one texts into some templates like ``\texttt{Because of \{1v1 text\}, \{class 1\} is different from \{class 2\}}''. 

\subsection{Use {\scshape{Coder}} on Downstream Tasks}
\begin{figure*}
  \centering
  \begin{subfigure}{0.82\linewidth}
    \includegraphics[width=\linewidth]{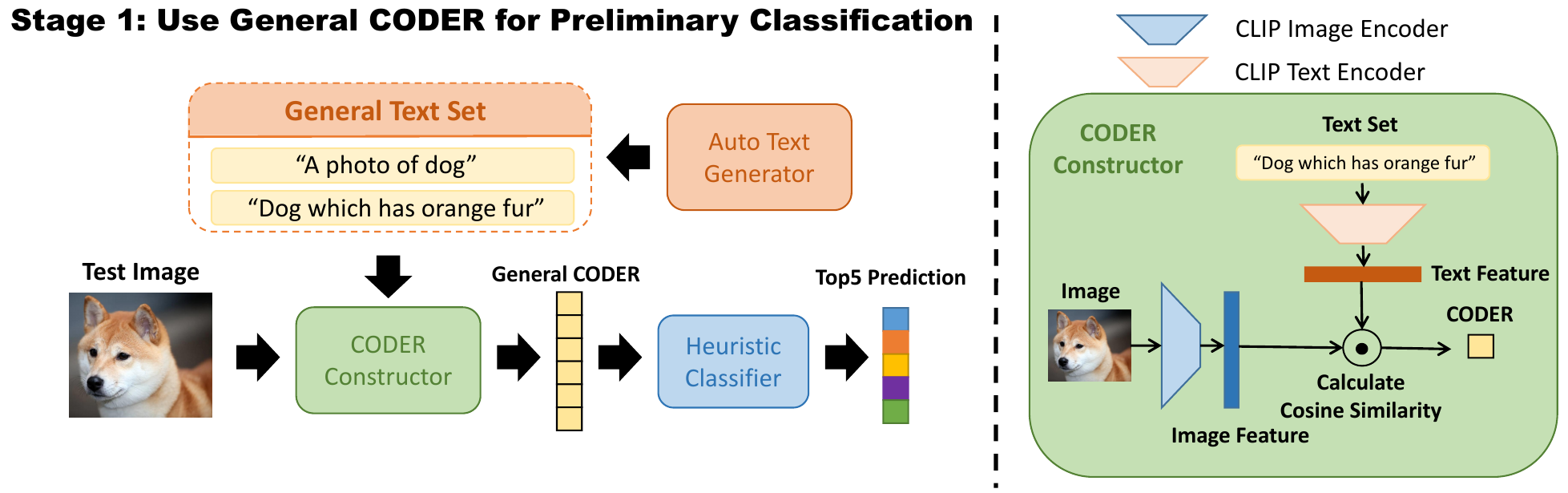}
    % \label{fig:short-a}
  \end{subfigure}
  \begin{subfigure}{0.82\linewidth}
    \includegraphics[width=\linewidth]{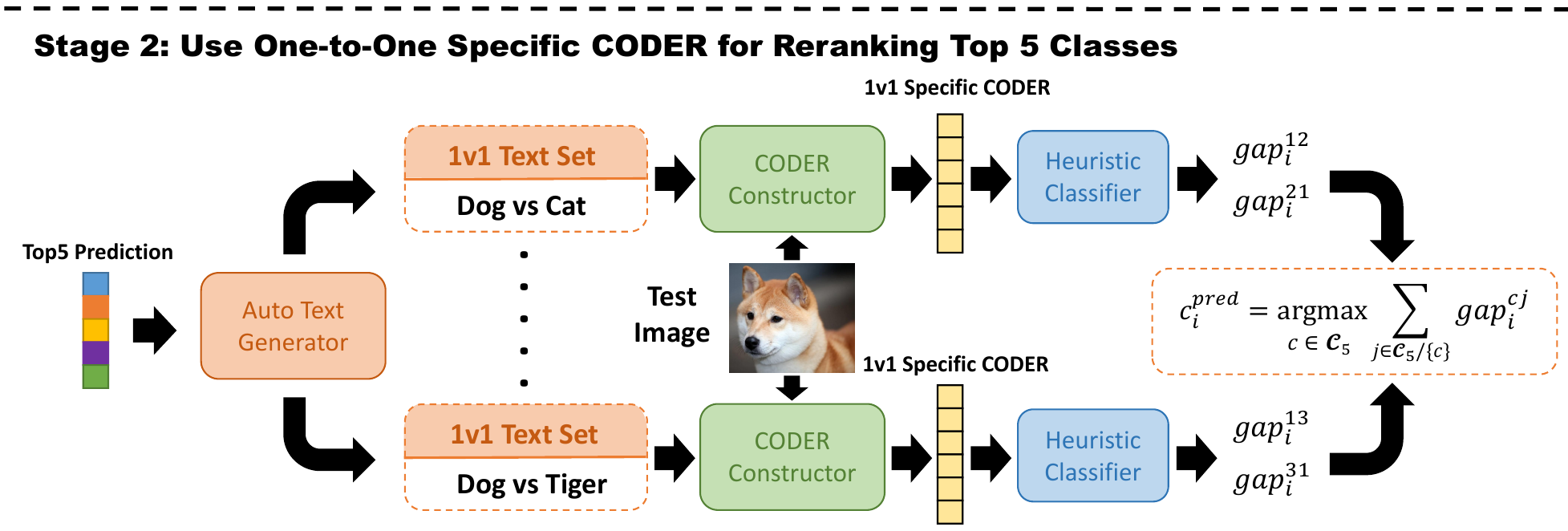} 
    % \label{fig:short-b}
  \end{subfigure}
  \caption{\textbf{Illustration of two-stage zero-shot image classification process based on image's \CODER.} In the first stage, we use the Auto Text Generator to create a General Text Set, which contains general descriptions of classes. This set is utilized to construct the image's general \CODER, and we use it for preliminary classification. In the second stage, we construct One-to-One Text Sets in pairs for the top five predicted classes of the preliminary classification results, focusing on attributes where the two specific classes differ most. We build one-to-one specific \CODER for the image based on these One-to-One Text Sets and use heuristic classifier to get each class's classification score. Then we rerank the top five preliminary results based on the classification score gaps $gap_i^{cj}$ between classes. Here $gap_i^{cj}$ represents the difference obtained by subtracting the score of the class $c$ from that of the class $j$ for image $\x_i$.}
  \label{rerank}
  \vspace{-0.35cm}
\end{figure*}
  
We apply \CODER in zero-shot and few-shot image classification tasks to enhance the performance of CLIP. 

\noindent{\bf Zero-Shot Image Classification.} In the first stage, we utilize a general text set generated by \ATG to construct the general \CODER for images. These texts can be considered as descriptions of general characteristics for a certain class in comparison to most other classes. We can derive preliminary classification results based on the general \CODER of the images. In the second stage, we employ \ATG to create distinct one-to-one specific text sets for the top five classes, each pair-wise. These texts serve as descriptions of the key distinguishing features between the classes. Utilizing these, we then develop corresponding one-to-one specific \CODER for images. Subsequently, we rerank the initial top five classification results using these one-to-one specific \CODER. \autoref{rerank} shows the process of our zero-shot image classification method. 

We first utilize all texts generated by \ATG, excluding the one-to-one texts, as the general text set $\mathcal{P}=\left[\vt_1, \cdots, \vt_{K}\right]$ to construct the general \CODER for test images. This process is shown in \autoref{zs1} and \autoref{zs2}. Here $\mW \in \bbR^{D_1 \times K}$ refers to the texts' features in the CLIP feature space. And $\vct{s}_i\in\bbR^{K}$ represents the \CODER of test image $\x_i$. $K$ refers to the total number of texts in $\mathcal{P}$.
\begin{equation}
\label{zs1}
    \mW = \left[\frac{\hat{\vt}_1}{\|\hat{\vt}_1\|}, \cdots, \frac{\hat{\vt}_{K}}{\|\hat{\vt}_{K}\|}\right]\in\mathbb{R}^{D_1 \times K}. 
\end{equation}
\begin{equation}
\label{zs2}
    \vct{s}_i = \phi\left(\hat{\x}_i\right) = \frac{\hat{\x}_i^{\top}}{\|\hat{\x}_i\|}\mW \in \mathbb{R}^{K}.
\end{equation}
Then, we employ a heuristic classifier $h$ on the test image's general \CODER $\vct{s}_i$ to obtain the preliminary classification logits vector $\bm{o}_i$. We use $\vct{s}_{ij}$ to represent the portion of the test image's general \CODER $\vct{s}_i$ corresponding to the text of the $j$-th category. $\vct{s}_{ij}$ consist of several parts, as shown in \autoref{zs3}. Here $\vct{s}_{ij}^{ori}$, $\vct{s}_{ij}^{att}$, $\vct{s}_{ij}^{ana}$, $\vct{s}_{ij}^{syn}$ refers to $\vct{s}_{ij}$'s portion of class name-based texts, attribute-based texts, analogous class-based texts and synonym-based texts, respectively. $\oplus$ refers to the vector concatenation operation. 
\begin{equation}
\label{zs3}
    \vct{s}_{ij} = \vct{s}_{ij}^{ori} \oplus \vct{s}_{ij}^{att} \oplus \vct{s}_{ij}^{ana} \oplus \vct{s}_{ij}^{syn}.
% \vspace{-0.6cm}
\end{equation}
For each class, the heuristic classifier $h$ first gets the largest element in the test image's \CODER portion corresponding to the class name-based texts and synonym-based texts. This step is intuitive: Humans can recognize an object by knowing just one of its names. Then $h$ calculates the mean of $\operatorname{max}(\vct{s}_{ij}^{ori} \oplus \vct{s}_{ij}^{syn})$ and all elements in the $\vct{s}_{ij}$'s portion corresponding to the attribute-based texts and synonym-based texts. This value can be seen as the preliminary classification logit value $o_{ij}$ belonging to class $j$. The process is shown in the following Equation:
\begin{equation}
\label{zs4}
    o_{ij} = h(\vct{s}_{ij}) = \operatorname{mean}\left(\vct{s}_{ij}^{att} \oplus \vct{s}_{ij}^{ana} \oplus \left[\operatorname{max}(\vct{s}_{ij}^{ori} \oplus \vct{s}_{ij}^{syn})\right]\right).
\end{equation}
After obtaining the preliminary classification results $\bm{o}_i$, we extract the top five classes from the prediction results, forming a set named $\mathcal{C}_{5}$. Subsequently, we draw all pairwise combinations from these five classes in set $\mathcal{C}_{5}$, creating a new set $\mathcal{P}_{1v1}$ comprised of these pairs. This process can be described by \autoref{eq-new1} and \autoref{eq-new2}:
\begin{equation}
\label{eq-new1}
    \mathcal{C}_{5} = \text{Top5}\left(\bm{o}_i\right)=\left\{c_{1}, c_{2}, c_{3}, c_{4}, c_{5}\right\}.
\end{equation}
\vspace{-0.3cm}
\begin{equation}
\label{eq-new2}
    \mathcal{P}_{1v1} = \left\{\left\{c_a, c_b\right\}|c_a,c_b\in \mathcal{C}_{5}, a\neq b\right\}.
\end{equation}

In the second stage, we use \ATG to generate the one-to-one specific texts for each pair in $\mathcal{P}_{1v1}$, resulting in a total of $C_5^2=10$ text sets. For each one-to-one specific text set, we can construct a one-to-one specific \CODER for the image. We first use CLIP's text encoder $f^{T}$ to extract the text features of the one-to-one text set $\mW^{a,b} \in \bbR^{D_1 \times (K_1+K_2)}$. Here $K_1$ and $K_2$ represent the number of one-to-one texts for each of the two classes. \autoref{eq-new3} shows the composition of $\mW^{a,b}$, where $\hat{\vt}_i^{a}$ and $\hat{\vt}_i^{b}$ refers to the $i$-th one-to-one text feature of class $a$ and class $b$, respectively.
\begin{equation}
\label{eq-new3}
    \mW^{a,b} = \left[\frac{\hat{\vt}_1^{a}}{\|\hat{\vt}_1^{a}\|}, \cdots, \frac{\hat{\vt}_{K_1}^{a}}{\|\hat{\vt}_{K_1}^{a}\|}, \frac{\hat{\vt}_{1}^{b}}{\|\hat{\vt}_{1}^{b}\|}, \cdots, \frac{\hat{\vt}_{K_2}^{b}}{\|\hat{\vt}_{K_2}^{b}\|}\right].
\end{equation}
Then we use the one-to-one text features $\mW^{a,b}$ to construct the image's one-to-one specific \CODER $\vct{s}_{i}^{a,b}\in\bbR^{K_1+K_2}$:
\begin{equation}
\label{eq-new4}
    \vct{s}_{i}^{a,b} = \frac{\hat{\x}_i^{\top}}{\|\hat{\x}_i\|}\mW^{a,b} = \vct{s}_{ia}^{a,b} \oplus \vct{s}_{ib}^{a,b}.
\end{equation}
Here $\vct{s}_{ia}^{a,b}\in\bbR^{K_1}$ and $\vct{s}_{ib}^{a,b}\in\bbR^{K_2}$ represent the part of \CODER pertaining to the one-to-one texts for class $a$ and class $b$, respectively.

Based on the one-to-one specific \CODER $\vct{s}_{i}^{a,b}$ of the image $\boldsymbol{x}_i$, we employ a heuristic classifier $h$ to obtain image's classification scores for class $a$ and $b$, denoted as $o_{ia}^{a,b}$ and $o_{ib}^{a,b}$, respectively:
\begin{equation}
\begin{aligned}
    o_{ia}^{a,b} = h(\vct{s}_{ia}^{a,b}) = \operatorname{mean}\left(\vct{s}_{ia}^{a,b}\right), \\
    o_{ib}^{a,b} = h(\vct{s}_{ib}^{a,b}) = \operatorname{mean}\left(\vct{s}_{ib}^{a,b}\right).
    \label{eq-new5}
\end{aligned}
\end{equation}
Then we calculate the image's classification score gap for each class relative to another class:
\begin{equation}
\label{eq-new6}
    gap_{i}^{ab} = o_{ia}^{a,b}-o_{ib}^{a,b}, \quad gap_{i}^{ba} = o_{ib}^{a,b}-o_{ia}^{a,b}.
\end{equation}
For each class, we sum up all the score gaps between it and other classes. Then we select the class with the largest sum of score gaps as the final predicted class $\hat{c}_{i}$ of the image:
\begin{equation}
\label{eq-new7}
    \hat{c}_{i}=\underset{c \in \mathcal{C}_5}{\operatorname{argmax}} \sum_{j \in \mathcal{C}_5 /\{c\}} gap_{i}^{cj}.
\end{equation}
We use the method based on score gaps instead of the traditional voting method to reorder initial prediction results. The reason is that score gaps provide quantified information about the relative advantages between classes. Unlike voting, which only considers each class's number of wins in pairwise classification, score gaps better capture the model's uncertainty in the classification tasks. For example, a small score gap might indicate that the model is uncertain about the correct class for the image, suggesting that the classification result may be unreliable. This can lead to errors in the outcomes derived from voting methods.

This two-phase image classification process illustrates that by changing the neighbor text set, we can dynamically build image's feature that focus on different semantics according to task requirements. 
In coarse-grained image classification, we focus on the general semantics of a class, such as whether an animal has wings. In fine-grained image classification, we aim to focus on more detailed features that differentiate between two specific classes, such as the color of wings when distinguishing between dragonfly and butterfly. However, the original image features extracted by the CLIP's image encoder cannot dynamically adapt to specific classification task requirements.
\begin{table*}[!htbp]
  \centering
  \caption{ Accuracy gains over VCD and CLIP baseline. The ``CODER'' column shows the classification accuracy before rerank stage, while the ``$\text{CODER}^{\star}$'' column shows the accuracy after rerank stage. $\Delta$ represents the improvement of our method over VCD.}
  \vspace{-0.3cm}
  \begin{adjustbox}{max width=\linewidth}
    \begin{tabular}{lccccccccccccccc}
    \addlinespace
    \toprule
     & \multicolumn{ 5}{c}{ImageNet} & \multicolumn{ 5}{c}{CUB200} & \multicolumn{ 5}{c}{EuroSAT} \\
    % \midrule
    \cmidrule(r){2-6}  \cmidrule(r){7-11} \cmidrule(r){12-16}
    Architecture     & CLIP & VCD & CODER & CODER$^{\star}$ & \textcolor{red}{$\Delta$} & CLIP & VCD & CODER & CODER$^{\star}$& \textcolor{red}{$\Delta$} & CLIP & VCD & CODER & CODER$^{\star}$ &  \textcolor{red}{$\Delta$}  \\
    ViT-B/32 & 59.07 & 62.96 & 64.38 & \textbf{66.86} &\textcolor{red}{\text{3.90}} & 51.81 & 51.78 & \text{53.00} & \textbf{55.04} & \textcolor{red}{3.26} & 44.89 & 46.57 &52.83& \textbf{54.93} & \textcolor{red}{8.36}\\
    ViT-B/16 & 63.53 & \text{68.10} & 69.61&\textbf{71.46} & \textcolor{red}{3.36} & 55.75 & 57.54 & \text{58.30}& \textbf{59.92} & \textcolor{red}{2.38} & 49.52 & 56.46 &\text{55.30}& \textbf{60.54} & \textcolor{red}{4.08}\\
    ViT-L/14 & 70.58 & 74.96 & 76.15&\textbf{77.38} & \textcolor{red}{2.42} & \text{62.10} & 63.29 & 64.64&\textbf{65.76} & \textcolor{red}{2.47} & 53.54 & 59.57 &62.98& \textbf{68.24}  &\textcolor{red}{8.67} \\
    ViT-L/14@336px & 71.81 & 76.11 & 77.27&\textbf{78.49} &\textcolor{red}{2.38} & 63.48 & 64.79 & 66.29 &\textbf{67.36} & \textcolor{red}{2.57} & 54.93 & 59.46 & 64.94&\textbf{69.96}  & \textcolor{red}{\text{10.50}}\\
    \toprule
     \toprule
          & \multicolumn{ 5}{c}{Describable Textures} & \multicolumn{ 5}{c}{Places365} & \multicolumn{ 5}{c}{Food101}  \\
          \cmidrule(r){2-6}  \cmidrule(r){7-11} \cmidrule(r){12-16}
    Architecture     & CLIP & VCD & CODER & CODER$^{\star}$ & \textcolor{red}{$\Delta$} & CLIP & VCD & CODER & CODER$^{\star}$& \textcolor{red}{$\Delta$} & CLIP & VCD & CODER & CODER$^{\star}$ &  \textcolor{red}{$\Delta$}  \\
    ViT-B/32 & 41.28 & 45.27 & 48.78 & \textbf{52.71} & \textcolor{red}{7.44} & 36.47 & 38.92 & \textbf{40.37} & 40.27 & \textcolor{red}{1.45} & 80.27 & 84.13 & 84.72&\textbf{85.50} & \textcolor{red}{1.37} \\
    ViT-B/16 & 43.19 & 45.64 &48.67& \textbf{55.69} &\textcolor{red}{10.05} & 37.31 & 39.87 &41.39& \textbf{\text{41.70}} &\textcolor{red}{1.83} & 85.99  & 89.25 & 89.29&\textbf{89.82} &\textcolor{red}{0.57}\\
    ViT-L/14 & 52.18 & 57.18 &60.74& \textbf{61.86} &\textcolor{red}{4.68}  & 37.06 & 38.55 & 41.21&\textbf{41.59} &\textcolor{red}{3.04} & 89.85 & 93.33 & 93.66&\textbf{93.93} &\textcolor{red}{\text{0.60}}\\
    ViT-L/14@336px & 52.12 & 57.45 & 61.12&\textbf{62.87} & \textcolor{red}{5.42} & 37.27 & 39.88 & 42.09&\textbf{\text{42.50}} &\textcolor{red}{2.62} & \text{91.10} & 94.06 & 94.48&\textbf{94.75} &\textcolor{red}{0.69}\\
    \toprule
     \toprule
          & \multicolumn{ 5}{c}{Caltech101} & \multicolumn{ 5}{c}{Oxford Pets} & \multicolumn{ 5}{c}{ImageNetV2}  \\
          \cmidrule(r){2-6}  \cmidrule(r){7-11} \cmidrule(r){12-16}
    Architecture     & CLIP & VCD & CODER & CODER$^{\star}$ & \textcolor{red}{$\Delta$} & CLIP & VCD & CODER & CODER$^{\star}$& \textcolor{red}{$\Delta$} & CLIP & VCD & CODER & CODER$^{\star}$ &  \textcolor{red}{$\Delta$}  \\
    ViT-B/32 & \text{79.90} & \text{89.80} & 91.24&\textbf{91.42} & \textcolor{red}{1.62} & 81.63 & 85.91 & 88.47&\textbf{89.26} &\textcolor{red}{3.35} & 51.79 & 55.29 & \textbf{56.75}&56.42 &\textcolor{red}{1.46}\\
    ViT-B/16 & 80.18 & 92.28  & \textbf{94.07}&93.95 &\textcolor{red}{1.79}  & 83.95 & 89.34 & \text{91.50}&\textbf{92.01} & \textcolor{red}{2.67} & 57.25 & 61.53 & 62.94&\textbf{62.97}  &\textcolor{red}{1.44}\\
    ViT-L/14 & 79.78 & 93.95 & \textbf{95.79} & 95.68 & \textcolor{red}{1.84} & \text{87.90} & 93.18 & 94.19&\textbf{94.36} & \textcolor{red}{1.18} & 64.31 & 69.21 & 70.35&\textbf{70.37} & \textcolor{red}{1.16} \\
    ViT-L/14@336px & 80.36 & 94.35 & 96.31&\textbf{96.37} & \textcolor{red}{2.02}& 87.82 & 93.62 & 94.22&\textbf{94.88} & \textcolor{red}{1.26}& 65.61 & 70.41 & \textbf{71.45}&71.28 & \textcolor{red}{1.04}\\
    \bottomrule
    \end{tabular}
    \end{adjustbox}
  \label{table1}
  \vspace{-0.2cm}
\end{table*}

\noindent{\bf Few-Shot Image Classification.} For Few-Shot Image Classification, we improve the Tip-Adapter~\citep{tip-adapter} by replacing the original CLIP's image features with our \CODER to calculate the similarity between the test image and the support set's images. We refer to the improved method as ~\textbf{C}r\textbf{O}ss-Mo\textbf{D}al N\textbf{E}ighbor \textbf{R}epresentation CLIP Adapter~(\CODER-Adapter). 
Given the $N$-way $M$-shot support image set $\mathcal{I}$, we first calculate the \CODER of the images in the support set, represented by $\bm{S}_{train}\in\bbR^{NM\times K}$. \autoref{fs1} shows the calculation process, where $f^{I}\left(\mathcal{I}\right)\in\bbR^{NM\times D_1}$ refers to the support images' original features and $\mW \in \bbR^{D_1 \times K}$ refers to the features of texts in the general text set $\mathcal{P}$, as defined in \autoref{zs1}.

\begin{equation}
\label{fs1}
    \bm{S}_{train} = f^{I}\left(\mathcal{I}\right)\mW.
\end{equation}
Then we perform one-hot encoding on the support images' labels $\bm{L}$ to get one-hot labels matrix $\bm{L}_{train}\in\bbR^{NM\times C}$.
Each row of $\bm{L}_{train}$ is a one-hot vector.
\begin{equation}
\label{fs2}
    \bm{L}_{train} = \operatorname{OneHot}\left(\bm{L}\right).
\end{equation}
For the test image $\x_{i}$, we also construct its \CODER $\bm{s}_i\in\bbR^{K}$ similarly to the support set's images.
\begin{equation}
\label{fs3}
    \vct{s}_i = f^{I}\left(\x_i\right)\mW.
\end{equation}
We then calculate the affinity matrix $\bm{A}\in\bbR^{1\times NM}$ between the test image's \CODER $\vct{s}_i\in\bbR^{K}$ and the support set images' \CODER $\bm{S}_{train}\in\bbR^{NM\times K}$ using the \autoref{fs4}. Here $\text{Norm}(\cdot)$ refers to the data normalization operation like L2 or min-max normalization. $\beta$ and $T$ are hyperparameters to control the sharpness of $\bm{A}$'s distribution. 
\begin{equation}
\label{fs4}
    \bm{A} = \exp\left(-\beta\cdot \left(1-\frac{\text{Norm}\left(\bm{s}_i \bm{S}_{train}^{\top}\right)}{T}\right)\right).
\end{equation}
The affinity $\bm{A}\in\bbR^{1\times NM}$ serves as a weight factor for $\bm{L}_{train}$. By calculating the weighted sum of images' labels in $\bm{L}_{train}$, we can refine the CLIP's original zero-shot prediction results $\bm{o}^{zs}_{i}$ using \autoref{fs5}. Here $\alpha$ controls the degree of correction.
\begin{equation}
\label{fs5}
    \bm{o}_{i} = \alpha \cdot \bm{A}\bm{L}_{train} + \bm{o}^{zs}_{i}.
\end{equation}
Finally, we select the class corresponding to the largest logits in $\bm{o}_{i}$ as the predicted class of image $\x_i$.
\section{Experiments}
\subsection{Zero-Shot Image Classification Performance}
For zero-shot image classification, we compare our method with two baselines: Vanilla CLIP~\citep{CLIP} and VCD~\citep{vcd}. 
Table \ref{table1} displays zero-shot image classification experiment results. By analyzing the results, we can draw the following conclusions: (1) Our proposed \CODER consistently boosts CLIP's zero-shot image classification accuracy across various datasets and model architectures; (2) The rerank stage based on the one-to-one specific \CODER can further enhance CLIP's performance, demonstrating the effectiveness of our proposed two-stage zero-shot classification method.

\subsection{Few-Shot Image Classification Performance}
For few-shot image classification, we use ResNet 50\cite{ResNet} as the backbone for CLIP's image encoder, consistent with previous methods. We use two CLIP's training-free few-shot image classification methods TIP-Adapter~\citep{tip-adapter} and TIP-X~\citep{sus-x} as our baselines. TIP-Adapter leverages features extracted by CLIP's image encoder for calculating the similarity between test images and support set's images. While TIP-X uses features based on similarity scores between images and texts generated by CUPL~\citep{cupl} to do it.

\begin{figure*}[htbp]
  \centering
  \begin{subfigure}{0.23\linewidth}
    \includegraphics[width=\linewidth]{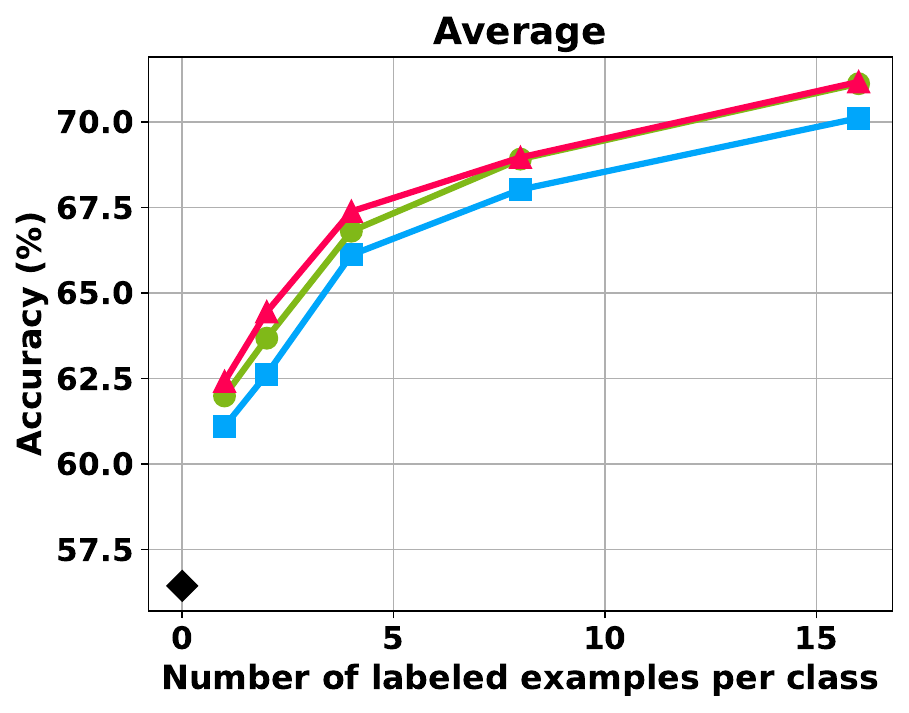}
   \end{subfigure}
  \begin{subfigure}{0.23\linewidth}
    \includegraphics[width=\linewidth]{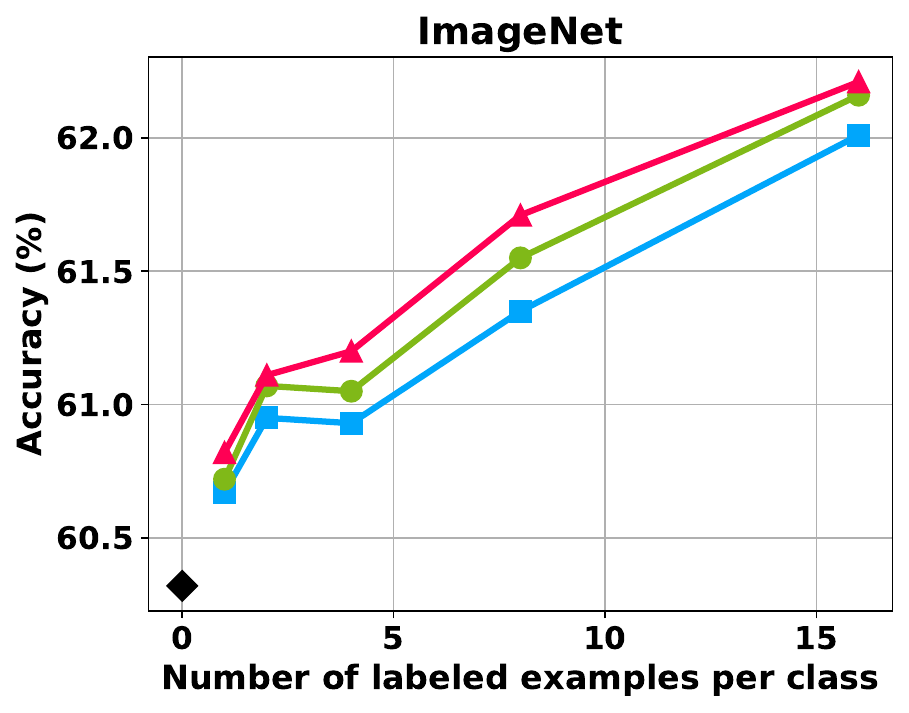} % 
  \end{subfigure}
    \begin{subfigure}{0.23\linewidth}
    \includegraphics[width=\linewidth]{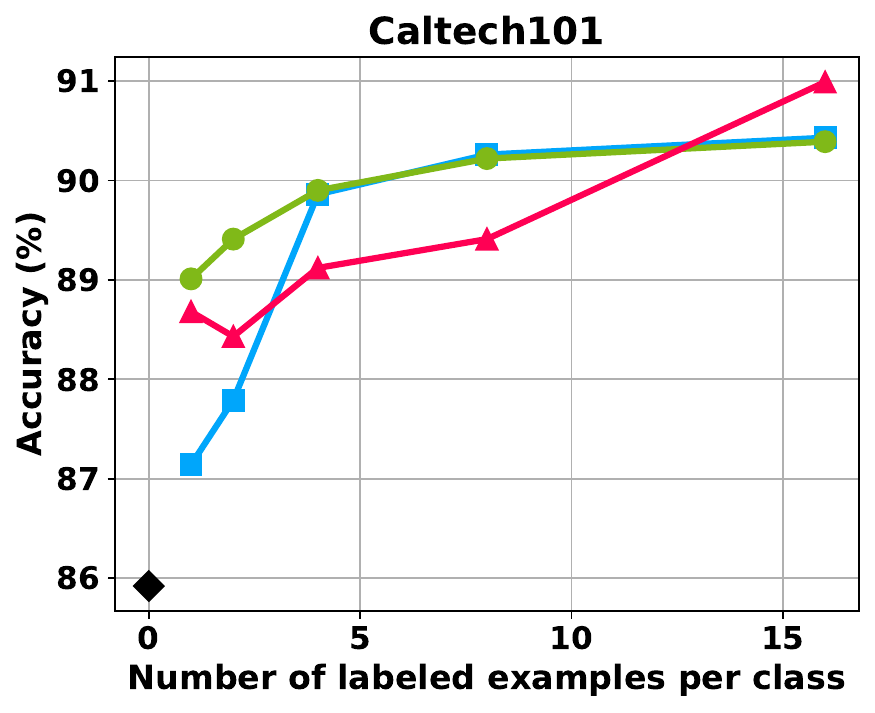} % 
  \end{subfigure}
  \begin{subfigure}{0.23\linewidth}
    \includegraphics[width=\linewidth]{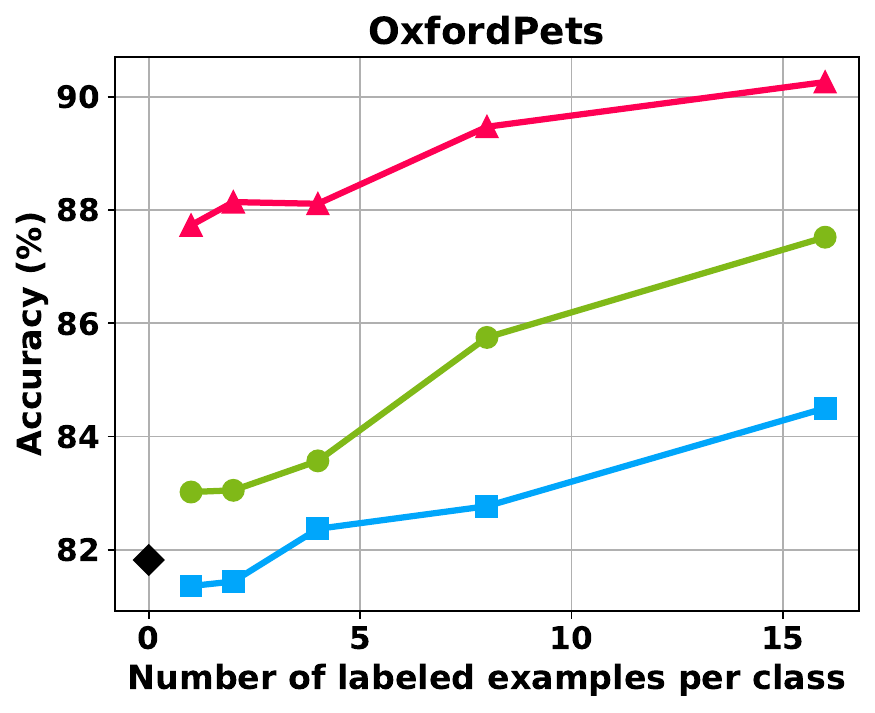} % 
  \end{subfigure}
  \begin{subfigure}{0.23\linewidth}
    \includegraphics[width=\linewidth]{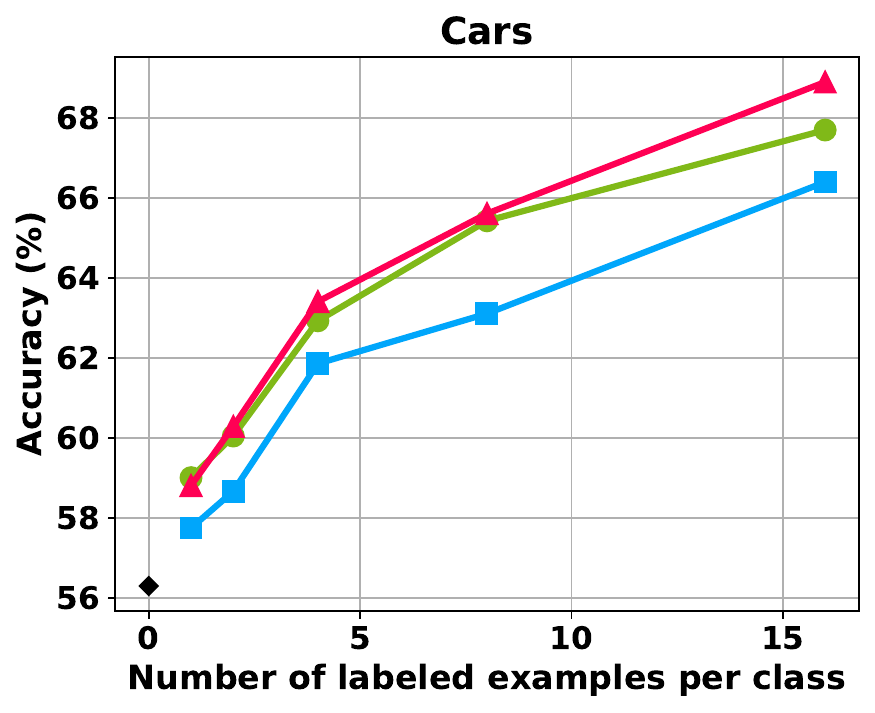} % 
  \end{subfigure}
  \begin{subfigure}{0.23\linewidth}
    \includegraphics[width=\linewidth]{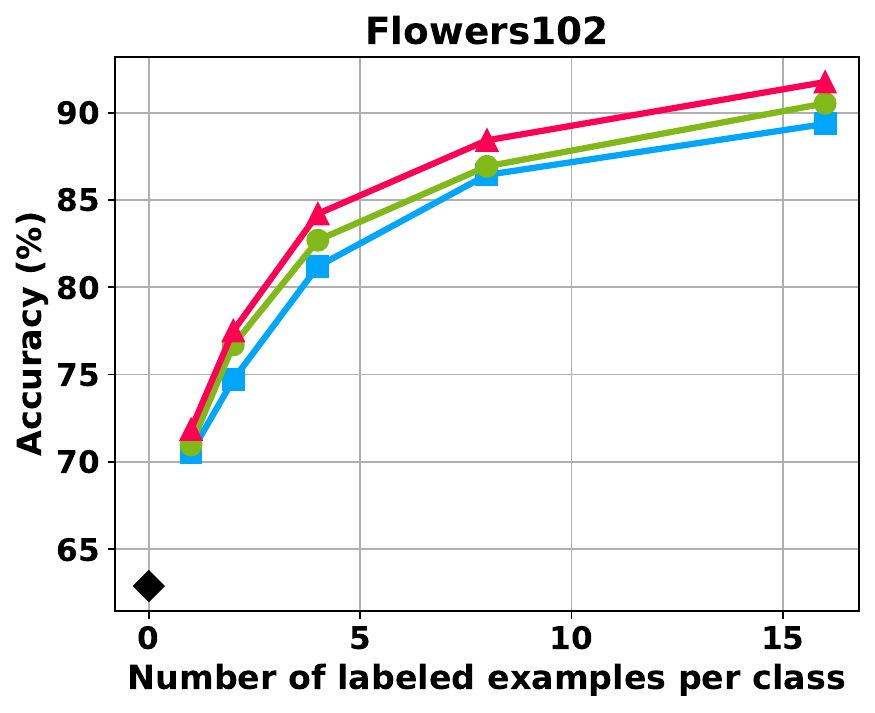} % 
  \end{subfigure}
  \begin{subfigure}{0.23\linewidth}
    \includegraphics[width=\linewidth]{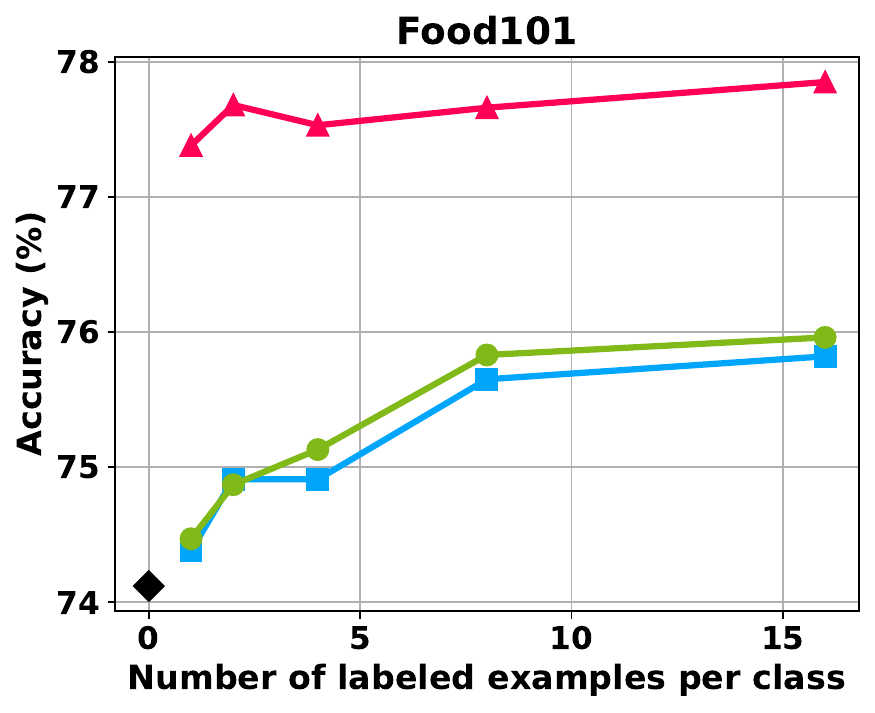} % 
  \end{subfigure}
  \begin{subfigure}{0.23\linewidth}
    \includegraphics[width=\linewidth]{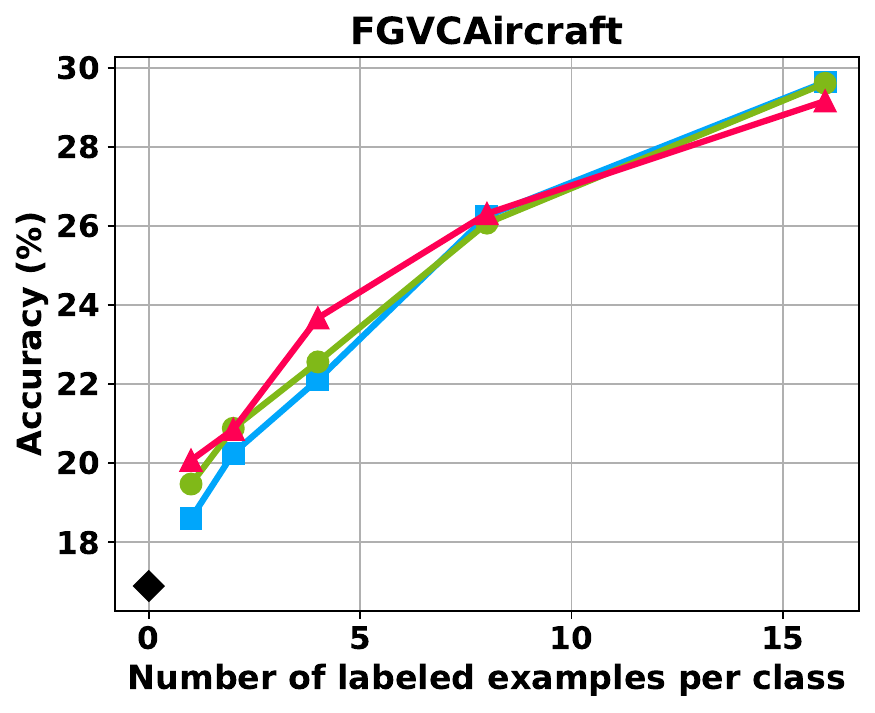} % 
  \end{subfigure}
  \begin{subfigure}{0.23\linewidth}
    \includegraphics[width=\linewidth]{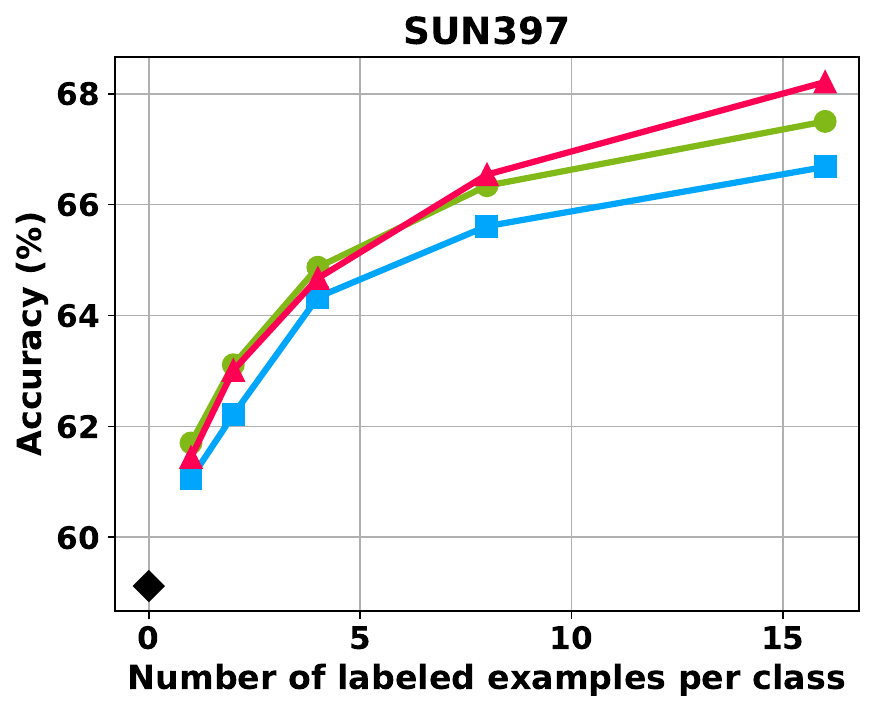} % 
  \end{subfigure}
    \begin{subfigure}{0.23\linewidth}
    \includegraphics[width=\linewidth]{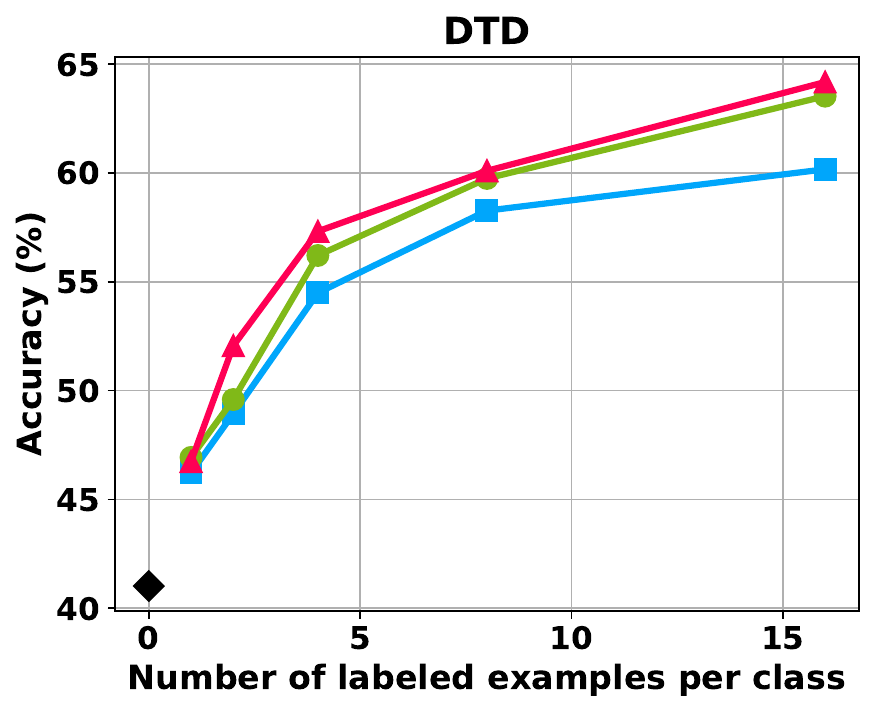} % 
  \end{subfigure}
    \begin{subfigure}{0.23\linewidth}
    \includegraphics[width=\linewidth]{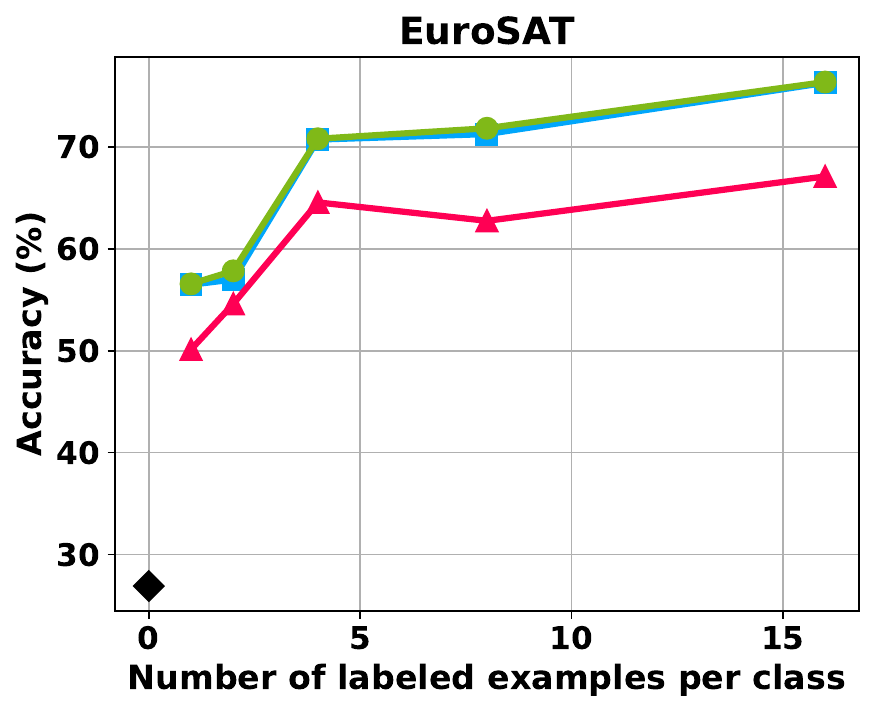} % 
  \end{subfigure}
  \begin{subfigure}{0.23\linewidth}
    \includegraphics[width=\linewidth]{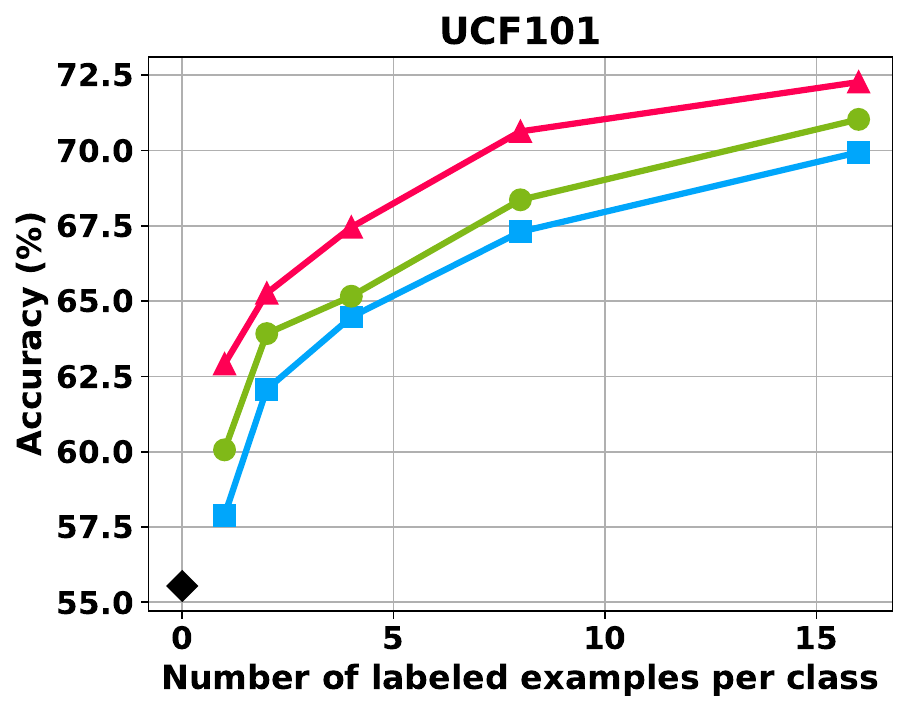} % 
  \end{subfigure}
  \begin{subfigure}{0.4\linewidth}
    \includegraphics[width=\linewidth]{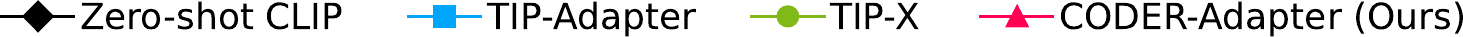} % 
  \end{subfigure}
  \caption{Results for the training-free few-shot regime across 11 datasets. We compare the \CODER-Adapter with the previous CLIP few-shot image classification methods. Our \CODER-Adapter achieves the best performance on most datasets.}
  \label{few-shot}
  \vspace{-0.3cm}
\end{figure*}

\autoref{few-shot} shows the few-shot image classification experiment results. We use the experimental results of TIP-Adapter and TIP-X presented in \cite{sus-x} and ensure that our random seeds are consistent with them. Our method surpasses the previous CLIP few-shot training-free image classification methods on most datasets across different shot scenarios. We notice that \CODER-Adapter's performance on EuroSAT is unsatisfactory, which results in our method not having a significant advantage in average accuracy across 11 datasets compared to previous methods. But this meets our expectations. Since EuroSAT has only 10 classes and 95 texts generated by \ATG, this small number of texts fails to satisfy \CODER's need for dense text sampling, impacting the adapter's performance. This highlights the importance of dense sampling for \CODER. 

Contrasting to previous work~\citep{tip-adapter,sus-x}, we innovatively interpret the superiority of CLIP's cross-modal relative feature representation from the perspective of neighbor representation. Additionally, by drawing an analogy to neighbor-based algorithms like $K$NN, which require dense sampling of neighbor samples to ensure superior performance, we provide an intuitive explanation for our need to use a more diverse and higher-quality text set to construct our \CODER. Motivated by this conclusion, we develop the Auto Text Generator for the automated creation of varied, high-quality texts to meet \CODER's dense sampling needs. The experiments demonstrate the validity of our method. 

\begin{table}[t]
  \centering
  \caption{ The accuracy of the zero-shot image classification experiments using different texts. Meaning of symbols: $\textbf{P}$: Using class name-based texts. $\textbf{Att}$: Using attribute-based texts. $\textbf{Ana}$: Using analogous class-based texts. We use CLIP ViT-B-32.}
  % \vspace{-0.2cm}
  %$\textbf{S}$: Using synonum-based texts. 
  \begin{adjustbox}{max width=\linewidth}
    \begin{tabular}{lcccc}
    \toprule 
    % \midrule
    \textbf{Text} & Caltech101 & Oxford Pets & Describable Textures &  EuroSAT   \\
        \cmidrule(r){1-5}
    P & 79.90 & 81.63 & 41.28 & 44.89 \\
    P+Att & 89.80 & 85.91 & 45.27 & 46.57 \\
    P+Att+Ana & \textbf{91.01} &\textbf{88.55} & \textbf{48.73} & \textbf{52.77}  \\
    \toprule
    \end{tabular}
    \end{adjustbox}
  \label{table4}
  \vspace{-0.2cm}
\end{table}
\begin{table}[t]
  \centering
  \caption{The accuracy of the 16-shot image classification experiments using different texts. We use CLIP ResNet-50.}
  % \vspace{-0.2cm}
  \begin{adjustbox}{max width=\linewidth}
    \begin{tabular}{lcccc}
    \toprule 
    % \midrule
    \textbf{Text} & Caltech101 & Oxford Pets & Describable Textures &  EuroSAT   \\
        \cmidrule(r){1-5}
    P & 90.34 & 89.83 & 62.35 & 58.44 \\
    P+Att & 90.47 & \textbf{90.3} & 63.06 & 65.36 \\
    P+Att+Ana & \textbf{90.99} & 90.26 & \textbf{64.18} & \textbf{67.11}  \\
    \toprule
    \end{tabular}
    \end{adjustbox}
  \label{table5}
  \vspace{-0.6cm}
\end{table}

\subsection{The Importance of Dense Sampling for \CODER}
Nearest neighbor algorithms like $K$NN depend on dense sampling for good performance. We view our \CODER as a cross-modal neighbor representation. Hence, as the diversity and quantity of high-quality cross-modal neighbor texts increase, the constructed \CODER should correspondingly improve. We validate this idea through experiments. 
We use three different text sets for constructing the image's \CODER: (1) CLIP’s original class name-based texts; (2) class name-based texts and VCD's attribute-based texts; (3) our analogous class-baed texts added to (2). Our experiments in zero-shot and few-shot image classification, as detailed in \autoref{table4} and \autoref{table5}, show that \CODER performance enhances with the increasing diversity and number of texts. This highlights the importance of dense neighbor text sampling in improving \CODER quality.
\begin{figure}[t]
\centering
 \begin{subfigure}{.48\linewidth}
    \includegraphics[width=\linewidth]{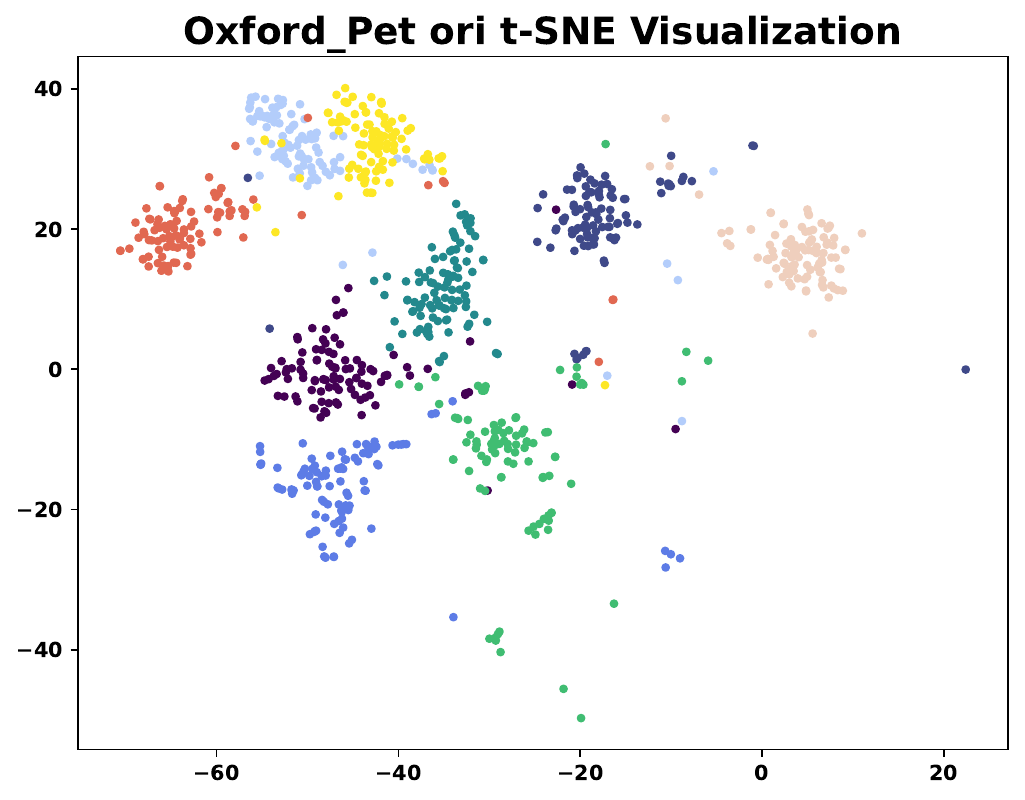} % 
  \end{subfigure}
\begin{subfigure}{.48\linewidth}
    \includegraphics[width=\linewidth]{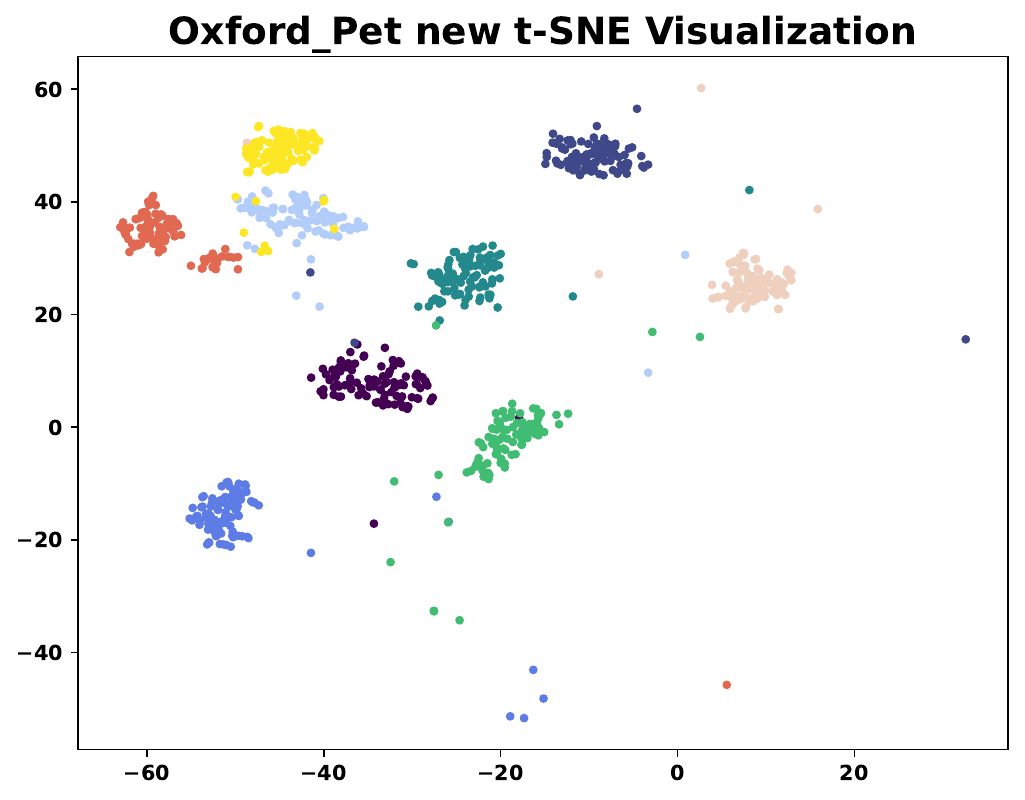} % 
  \end{subfigure}
  \caption{ The t-SNE Visualization of the Oxford-Pets dataset. \textbf{Left}: Original CLIP Images' Features; \textbf{Right}: \CODER.}
\label{figure5}
\vspace{-0.6cm}
\end{figure}

\subsection{Visualization results of \CODER}
\autoref{figure5} presents a t-SNE visualization comparing original image features from the CLIP image encoder with those of our proposed \CODER on the Oxford-Pets dataset. Our \CODER achieves tighter clustering of same-class images' features and clearer separation of different classes.

\section{Conclusion}
In this paper, we address the misalignment between CLIP's image feature extraction method and its pre-training paradigm. We present a novel perspective based on the nearest neighbors to comprehend CLIP's strong zero-shot image classification capability. Our key insight is that CLIP's effective text-image matching capability embeds image information in image-text distances. This leads us to propose the \textbf{C}r\textbf{O}ss-mo\textbf{D}al n\textbf{E}ighbor \textbf{R}epresentation~(\CODER), utilizing these image-text distances for image representation. We introduce the \textbf{A}uto \textbf{T}ext \textbf{G}enerator to automatically generate texts, ensuring dense sampling of neighbor texts for better \CODER construction. Experiment results in zero-shot and few-shot image classification show the superiority of our method.
\section*{Acknowledgments}
This work is partially supported by National Key R\&D Program of China (2022ZD0114805),
NSFC (62376118, 62006112, 62250069, 61921006), Collaborative Innovation Center of Novel Software Technology and Industrialization.

{
    \small
    \bibliographystyle{ieeenat_fullname}
    \bibliography{main}

\begin{thebibliography}{35}
\providecommand{\natexlab}[1]{#1}
\providecommand{\url}[1]{\texttt{#1}}
\expandafter\ifx\csname urlstyle\endcsname\relax
  \providecommand{\doi}[1]{doi: #1}\else
  \providecommand{\doi}{doi: \begingroup \urlstyle{rm}\Url}\fi

\bibitem[Fan et~al.(2023)Fan, Krishnan, Isola, Katabi, and Tian]{rewrite_clip}
Lijie Fan, Dilip Krishnan, Phillip Isola, Dina Katabi, and Yonglong Tian.
\newblock Improving {CLIP} training with language rewrites.
\newblock In \emph{NeurIPS}, 2023.

\bibitem[Ge et~al.(2023)Ge, Ren, Gallagher, Wang, Yang, Adam, Itti, Lakshminarayanan, and Zhao]{bottom-up}
Yunhao Ge, Jie Ren, Andrew Gallagher, Yuxiao Wang, Ming-Hsuan Yang, Hartwig Adam, Laurent Itti, Balaji Lakshminarayanan, and Jiaping Zhao.
\newblock Improving zero-shot generalization and robustness of multi-modal models.
\newblock In \emph{CVPR}, pages 11093--11101, 2023.

\bibitem[Gu et~al.(2022)Gu, Lin, Kuo, and Cui]{obj1}
Xiuye Gu, Tsung-Yi Lin, Weicheng Kuo, and Yin Cui.
\newblock Open-vocabulary object detection via vision and language knowledge distillation.
\newblock In \emph{ICLR}, 2022.

\bibitem[Guzhov et~al.(2022)Guzhov, Raue, Hees, and Dengel]{audioclip}
Andrey Guzhov, Federico Raue, J{\"{o}}rn Hees, and Andreas Dengel.
\newblock Audioclip: Extending clip to image, text and audio.
\newblock In \emph{ICASSP}, pages 976--980, 2022.

\bibitem[He et~al.(2016)He, Zhang, Ren, and Sun]{ResNet}
Kaiming He, Xiangyu Zhang, Shaoqing Ren, and Jian Sun.
\newblock Deep residual learning for image recognition.
\newblock In \emph{CVPR}, pages 770--778, 2016.

\bibitem[Jia et~al.(2021)Jia, Yang, Xia, Chen, Parekh, Pham, Le, Sung, Li, and Duerig]{align}
Chao Jia, Yinfei Yang, Ye Xia, Yi-Ting Chen, Zarana Parekh, Hieu Pham, Quoc~V. Le, Yun-Hsuan Sung, Zhen Li, and Tom Duerig.
\newblock Scaling up visual and vision-language representation learning with noisy text supervision.
\newblock In \emph{ICML}, pages 4904--4916, 2021.

\bibitem[Li et~al.(2022)Li, Weinberger, Belongie, Koltun, and Ranftl]{seg1}
Boyi Li, Kilian~Q. Weinberger, Serge~J. Belongie, Vladlen Koltun, and Ren{\'{e}} Ranftl.
\newblock Language-driven semantic segmentation.
\newblock In \emph{ICLR}, 2022.

\bibitem[Li et~al.(2021)Li, Selvaraju, Gotmare, Joty, Xiong, and Hoi]{albef}
Junnan Li, Ramprasaath~R. Selvaraju, Akhilesh Gotmare, Shafiq~R. Joty, Caiming Xiong, and Steven Chu-Hong Hoi.
\newblock Align before fuse: Vision and language representation learning with momentum distillation.
\newblock In \emph{NeurIPS}, pages 9694--9705, 2021.

\bibitem[Li et~al.(2023)Li, Peng, Chen, Gao, and Yang]{llm1}
Li Li, Jiawei Peng, Huiyi Chen, Chongyang Gao, and Xu Yang.
\newblock How to configure good in-context sequence for visual question answering.
\newblock \emph{CoRR}, abs/2312.01571, 2023.

\bibitem[Luo et~al.(2022)Luo, Ji, Zhong, Chen, Lei, Duan, and Li]{clip4clip}
Huaishao Luo, Lei Ji, Ming Zhong, Yang Chen, Wen Lei, Nan Duan, and Tianrui Li.
\newblock Clip4clip: An empirical study of {CLIP} for end to end video clip retrieval and captioning.
\newblock \emph{Neurocomputing}, 508:\penalty0 293--304, 2022.

\bibitem[Mao et~al.(2023)Mao, Teotia, Sundar, Menon, Yang, Wang, and Vondrick]{double_dought}
Chengzhi Mao, Revant Teotia, Amrutha Sundar, Sachit Menon, Junfeng Yang, Xin Wang, and Carl Vondrick.
\newblock Doubly right object recognition: {A} why prompt for visual rationales.
\newblock In \emph{CVPR}, pages 2722--2732, 2023.

\bibitem[Menon and Vondrick(2023)]{vcd}
Sachit Menon and Carl Vondrick.
\newblock Visual classification via description from large language models.
\newblock In \emph{ICLR}, 2023.

\bibitem[Ouyang et~al.(2022)Ouyang, Wu, Jiang, Almeida, Wainwright, Mishkin, Zhang, Agarwal, Slama, Ray, Schulman, Hilton, Kelton, Miller, Simens, Askell, Welinder, Christiano, Leike, and Lowe]{InstructGPT}
Long Ouyang, Jeffrey Wu, Xu Jiang, Diogo Almeida, Carroll~L. Wainwright, Pamela Mishkin, Chong Zhang, Sandhini Agarwal, Katarina Slama, Alex Ray, John Schulman, Jacob Hilton, Fraser Kelton, Luke Miller, Maddie Simens, Amanda Askell, Peter Welinder, Paul~F. Christiano, Jan Leike, and Ryan Lowe.
\newblock Training language models to follow instructions with human feedback.
\newblock In \emph{NeurIPS}, pages 27730--27744, 2022.

\bibitem[Peng et~al.(2023)Peng, Yang, Ma, Xu, Zhang, Han, and Zhang]{llm2}
Yingzhe Peng, Xu Yang, Haoxuan Ma, Shuo Xu, Chi Zhang, Yucheng Han, and Hanwang Zhang.
\newblock {ICD-LM:} configuring vision-language in-context demonstrations by language modeling.
\newblock \emph{CoRR}, abs/2312.10104, 2023.

\bibitem[Pratt et~al.(2022)Pratt, Liu, and Farhadi]{cupl}
Sarah~M. Pratt, Rosanne Liu, and Ali Farhadi.
\newblock What does a platypus look like? generating customized prompts for zero-shot image classification.
\newblock \emph{CoRR}, abs/2209.03320, 2022.

\bibitem[Radford et~al.(2021)Radford, Kim, Hallacy, Ramesh, Goh, Agarwal, Sastry, Askell, Mishkin, Clark, Krueger, and Sutskever]{CLIP}
Alec Radford, Jong~Wook Kim, Chris Hallacy, Aditya Ramesh, Gabriel Goh, Sandhini Agarwal, Girish Sastry, Amanda Askell, Pamela Mishkin, Jack Clark, Gretchen Krueger, and Ilya Sutskever.
\newblock Learning transferable visual models from natural language supervision.
\newblock In \emph{ICML}, pages 8748--8763, 2021.

\bibitem[Shen et~al.(2022)Shen, Li, Tan, Bansal, Rohrbach, Chang, Yao, and Keutzer]{clip-vil}
Sheng Shen, Liunian~Harold Li, Hao Tan, Mohit Bansal, Anna Rohrbach, Kai-Wei Chang, Zhewei Yao, and Kurt Keutzer.
\newblock How much can {CLIP} benefit vision-and-language tasks?
\newblock In \emph{ICLR}, 2022.

\bibitem[Singh et~al.(2022)Singh, Hu, Goswami, Couairon, Galuba, Rohrbach, and Kiela]{flava}
Amanpreet Singh, Ronghang Hu, Vedanuj Goswami, Guillaume Couairon, Wojciech Galuba, Marcus Rohrbach, and Douwe Kiela.
\newblock {FLAVA:} {A} foundational language and vision alignment model.
\newblock In \emph{CVPR}, pages 15617--15629, 2022.

\bibitem[Tao et~al.(2023)Tao, Bao, Tang, and Xu]{t2i2}
Ming Tao, Bing-Kun Bao, Hao Tang, and Changsheng Xu.
\newblock {GALIP:} generative adversarial clips for text-to-image synthesis.
\newblock In \emph{CVPR}, pages 14214--14223, 2023.

\bibitem[Udandarao et~al.(2022)Udandarao, Gupta, and Albanie]{sus-x}
Vishaal Udandarao, Ankush Gupta, and Samuel Albanie.
\newblock Sus-x: Training-free name-only transfer of vision-language models.
\newblock \emph{CoRR}, abs/2211.16198, 2022.

\bibitem[Vinker et~al.(2022)Vinker, Pajouheshgar, Bo, Bachmann, Bermano, Cohen-Or, Zamir, and Shamir]{clipasso}
Yael Vinker, Ehsan Pajouheshgar, Jessica~Y. Bo, Roman~Christian Bachmann, Amit~Haim Bermano, Daniel Cohen-Or, Amir Zamir, and Ariel Shamir.
\newblock Clipasso: semantically-aware object sketching.
\newblock \emph{ACM Transactions on Graphics}, 41\penalty0 (4):\penalty0 86:1--86:11, 2022.

\bibitem[Wang et~al.(2023)Wang, Bao, Dong, Bjorck, Peng, Liu, Aggarwal, Mohammed, Singhal, Som, and Wei]{beit-3}
Wenhui Wang, Hangbo Bao, Li Dong, Johan Bjorck, Zhiliang Peng, Qiang Liu, Kriti Aggarwal, Owais~Khan Mohammed, Saksham Singhal, Subhojit Som, and Furu Wei.
\newblock Image as a foreign language: {BEIT} pretraining for vision and vision-language tasks.
\newblock In \emph{CVPR}, pages 19175--19186, 2023.

\bibitem[Wei et~al.(2022)Wei, Xie, Zhou, Li, and Tian]{mvp}
Longhui Wei, Lingxi Xie, Wengang Zhou, Houqiang Li, and Qi Tian.
\newblock {MVP:} multimodality-guided visual pre-training.
\newblock In \emph{ECCV}, pages 337--353, 2022.

\bibitem[Wu(2011)]{wjx2}
Jianxin Wu.
\newblock Balance support vector machines locally using the structural similarity kernel.
\newblock In \emph{PAKDD}, pages 112--123, 2011.

\bibitem[Xu et~al.(2022)Xu, Mello, Liu, Byeon, Breuel, Kautz, and Wang]{seg2}
Jiarui Xu, Shalini~De Mello, Sifei Liu, Wonmin Byeon, Thomas~M. Breuel, Jan Kautz, and Xiaolong Wang.
\newblock Groupvit: Semantic segmentation emerges from text supervision.
\newblock In \emph{CVPR}, 2022.

\bibitem[Yan et~al.(2023)Yan, Wang, Zhong, Dong, He, Lu, Wang, Shang, and McAuley]{cbm2}
An Yan, Yu Wang, Yiwu Zhong, Chengyu Dong, Zexue He, Yujie Lu, William~Yang Wang, Jingbo Shang, and Julian~J. McAuley.
\newblock Learning concise and descriptive attributes for visual recognition.
\newblock In \emph{ICCV}, pages 3067--3077. {IEEE}, 2023.

\bibitem[Yang et~al.(2023{\natexlab{a}})Yang, Wu, Yang, Chen, and Geng]{llm3}
Xu Yang, Yongliang Wu, Mingzhuo Yang, Haokun Chen, and Xin Geng.
\newblock Exploring diverse in-context configurations for image captioning.
\newblock In \emph{NeurIPS}, 2023{\natexlab{a}}.

\bibitem[Yang et~al.(2023{\natexlab{b}})Yang, Panagopoulou, Zhou, Jin, Callison-Burch, and Yatskar]{cbm1}
Yue Yang, Artemis Panagopoulou, Shenghao Zhou, Daniel Jin, Chris Callison-Burch, and Mark Yatskar.
\newblock Language in a bottle: Language model guided concept bottlenecks for interpretable image classification.
\newblock In \emph{CVPR}, pages 19187--19197, 2023{\natexlab{b}}.

\bibitem[Yi et~al.(2024)Yi, De-Chuan, and Han-Jia]{select_vlm}
Chao Yi, Zhan De-Chuan, and Ye Han-Jia.
\newblock Bridge the modality and capacity gaps in vision-language model selection.
\newblock \emph{CoRR}, abs/2403.13797, 2024.

\bibitem[Yuan et~al.(2021)Yuan, Chen, Chen, Codella, Dai, Gao, Hu, Huang, Li, Li, Liu, Liu, Liu, Lu, Shi, Wang, Wang, Xiao, Xiao, Yang, Zeng, Zhou, and Zhang]{florence}
Lu Yuan, Dongdong Chen, Yi-Ling Chen, Noel Codella, Xiyang Dai, Jianfeng Gao, Houdong Hu, Xuedong Huang, Boxin Li, Chunyuan Li, Ce Liu, Mengchen Liu, Zicheng Liu, Yumao Lu, Yu Shi, Lijuan Wang, Jianfeng Wang, Bin Xiao, Zhen Xiao, Jianwei Yang, Michael Zeng, Luowei Zhou, and Pengchuan Zhang.
\newblock Florence: {A} new foundation model for computer vision.
\newblock \emph{CoRR}, 2021.

\bibitem[Zhang et~al.(2022{\natexlab{a}})Zhang, Guo, Zhang, Li, Miao, Cui, Qiao, Gao, and Li]{pointclip}
Renrui Zhang, Ziyu Guo, Wei Zhang, Kunchang Li, Xupeng Miao, Bin Cui, Yu Qiao, Peng Gao, and Hongsheng Li.
\newblock Pointclip: Point cloud understanding by {CLIP}.
\newblock In \emph{CVPR}, pages 8542--8552, 2022{\natexlab{a}}.

\bibitem[Zhang et~al.(2022{\natexlab{b}})Zhang, Zhang, Fang, Gao, Li, Dai, Qiao, and Li]{tip-adapter}
Renrui Zhang, Wei Zhang, Rongyao Fang, Peng Gao, Kunchang Li, Jifeng Dai, Yu Qiao, and Hongsheng Li.
\newblock Tip-adapter: Training-free adaption of {CLIP} for few-shot classification.
\newblock In \emph{ECCV}, pages 493--510, 2022{\natexlab{b}}.

\bibitem[Zhou et~al.(2015)Zhou, Wu, and song Zhou]{wjx1}
Guobing Zhou, Jianxin Wu, and song Zhou.
\newblock A nearest-neighbor-based clustering method.
\newblock \emph{Journal of Software}, 26\penalty0 (11):\penalty0 2847--2855, 2015.

\bibitem[Zhou et~al.(2022)Zhou, Zhang, Chen, Li, Tensmeyer, Yu, Gu, Xu, and Sun]{t2i1}
Yufan Zhou, Ruiyi Zhang, Changyou Chen, Chunyuan Li, Chris Tensmeyer, Tong Yu, Jiuxiang Gu, Jinhui Xu, and Tong Sun.
\newblock Towards language-free training for text-to-image generation.
\newblock In \emph{CVPR}, pages 17886--17896, 2022.

\bibitem[Zohar et~al.(2023)Zohar, Huang, Wang, and Yeung]{LOVM}
Orr Zohar, Shih-Cheng Huang, Kuan-Chieh Wang, and Serena Yeung.
\newblock Lovm: Language-only vision model selection.
\newblock In \emph{NeurIPS}, 2023.

\end{thebibliography}
}

\setcounter{section}{0}
\renewcommand{\thesection}{\Roman{section}}

%%%%%%%%%% Merge with supplemental materials %%%%%%%%%%
%%%%%%%%%% Prefix a "S" to all equations, figures, tables and reset the counter %%%%%%%%%%
\setcounter{equation}{0}
\setcounter{figure}{0}
\setcounter{table}{0}
\setcounter{page}{1}
\makeatletter
\clearpage
\setcounter{page}{1}
\maketitlesupplementary
\section{Analysis of Different Types of Texts}
In our proposed \CODER, each element of it can be regarded as the image's matching score with corresponding semantics. Therefore, in the following content, we will explain each text from the perspective of semantics to illustrate the advantages of the new text types we propose. Our Auto Text Generator~(\ATG) introduces three new types of semantics compared to previous works\citep{CLIP,vcd}: Analogous Class-based Semantics, Synonym-based Semantics, and One-to-One Specific Semantics. \autoref{table:images} shows examples of how our newly proposed semantics modify the original wrong classification results of images. For ease of demonstration, we only display partial results of the image's matching scores with the original semantics and the new semantics. Next, we will analyze the functions of these semantics one by one.

\subsection{The Role of Different Types of Semantics}
\textbf{Analogous Class-based Semantics.} Introducing analogous classes allows the model to utilize the knowledge of the analogous classes and the relationships between classes in image classification. This mimics the human process of identifying objects by comparing the similarity of unknown classes to known classes. Samples 1 to 5 in \autoref{table:images} demonstrate the correction of the images' original predicted classes by the analogous class-based semantics. 
Samples 1 and 2 show a side view of a biplane where the propeller is not visible, yet the landing gear is clearly visible.
Since the landing gear of most modern airplanes is retracted during flight, the attributes ChatGPT generates for airplanes do not include landing gear. In contrast, for helicopters, the landing gear is generally fixed and conspicuous, so its ChatGPT-generated attributes include landing gear. This leads to the error.
After we use the analogous class-based semantics of ground-truth class, we get semantics that more closely resembling the object in the image and then successfully correct the result. Similarly, in Sample 5, motion blur in the image led to a mistaken match with a spinning ceiling fan. By using analogous class-based semantics, we compare the similarity of the object in the image with both ``Floor Lamp'' and ``Industrial Fan''. Since the object in the picture is more similar to ``Floor Lamp'', we correct the results accordingly.

\noindent\textbf{Synonym-based Semantics.} Samples 6 to 9 in \autoref{table:images} demonstrate the correction of samples by the synonym semantics. These samples show that different synonyms for a class result in completely different matching scores for the same test image. 
By introducing multiple synonym semantics and selecting the maximum score among these, we effectively address the classification errors caused by CLIP's varying responses to different synonyms for the same class.

\noindent\textbf{One-to-One Specific Semantics.} Samples 10 to 12 in \autoref{table:images} showcase the correction of samples by the one-to-one specific semantics. Observing these samples, we notice that when the actual class of an image is very similar to the incorrectly predicted class by CLIP, these two classes often share many similar semantics. This leads to difficulties for the original semantics proposed in CLIP and VCD to distinguish between these classes, resulting in classification errors. For instance, in Sample 10, both the Staffordshire Bull Terrier and the Boxer share the semantic of short hair; in Sample 11, Siamese and Birman cats both have blue eyes. Our proposed one-to-one specific semantics leverage ChatGPT's knowledge to generate key distinguishing features for these specific classes. For example, in Sample 10, the one-to-one specific semantics focus on the difference in the underbite between Staffordshire Bull Terriers and Boxers. For Sample 11, ChatGPT notes the difference in fur length between Siamese and Birman Cats. 
For Sample 12, the one-to-one specific semantics highlight the flat-faced feature of Persian Cats.
These semantics are crucial for differentiating between the two similar classes, thereby helping to correct the image's original predicted class.

\subsection{Understanding Failure Cases}
\label{failure_case}
We also analyze some common errors of our method.

\noindent\textbf{Analogous Class-based Semantics.} Regarding the analogous class-based semantics, we have identified five common types of errors:
\begin{enumerate}
    \item The analogous class-based semantics struggle when the analogous class of the incorrectly predicted class closely resembles the true class. In Sample 1 of \autoref{table:images2}, the analogous class for emu is Cassowary, and for llama is Camel. 
    The emu in the image is misidentified as a camel due to its color resembling that of a camel and being set against a desert-like background.
    \item When class names are ambiguous, the analogous class-based semantics may incorrectly associate them with a similar class that doesn't match the image due to the ambiguous meaning of the word. For example, in Sample 2 of \autoref{table:images2}, ``bass'' refers to both a fish and a musical instrument. Therefore, \ATG generates a wrong analogous class, ``upright bass'', that doesn't match the image. To avoid such errors, it's essential to clarify class names, like changing ``bass'' to ``bass fish'' to remove ambiguity.
    \item When the object in an image lacks the common semantics of its class, it struggles to match with its analogous classes. As shown in Sample 3 of \autoref{table:images2}, the crab in the image doesn't prominently display typical characteristics, such as two large claws and eight legs, making it difficult to have a high matching score with its analogous classes like King Crab.
    \item When the analogous classes of an incorrectly predicted class also appear in the image, it can lead to error. For example, in Sample 4 of \autoref{table:images2}, the analogous class for Sea Horse is Aquarium Decorations and the presence of Aquarium Decorations in the image results in a misclassification.
    \item When the label of the image itself is ambiguous, the analogous class-based semantics may also encounter issues. As demonstrated in Sample 5 of \autoref{table:images2}, the image shows an artwork shaped like a dragonfly made from leaves. It's unclear whether to assign the label of a dragonfly or a type of plant to this image, leading to a high matching score with analogous classes like desert plant.
\end{enumerate}
\vspace{10pt}

\noindent\textbf{Synonym-based Semantics.} For synonym-based semantics, when a class name has multiple meanings, its synonyms may correspond to a different meaning that doesn't match the current image, leading to incorrect classification results. For example, in Sample 6 of \autoref{table:images2}, "Bombay" and "Siamese" refer to both cat breeds and place or cultural names, resulting in identified synonyms that may not be cat breeds. This issue can be resolved by clarifying class names to eliminate ambiguity, such as renaming ``Bombay'' to ``Bombay Cat''.
\vspace{10pt}

\noindent\textbf{One-to-One Specific Semantics.} Regarding the one-to-one specific semantics, we have identified two types of errors:
\begin{enumerate}
    \item \textbf{The Bottleneck in CLIP's Recognition Ability.} Although the Auto Text Generator~(\ATG) produces texts highlighting key semantics to differentiate between two classes, CLIP's image-text matching capability may have limitations in certain image-text pairs, preventing the one-to-one specific semantics from producing accurate results. For instance, in Sample 7 of \autoref{table:images2}, \ATG identifies that a Bengal Cat has a marbled coat. However, the matching score of the image and the ``marbled coat'' semantic is low, even though the cat in the image indeed displays a marbled coat. This issue arises from CLIP's inability to precisely calculate the similarity between certain image-text pairs.
    \item \textbf{The Capacity Limitations of External Experts like ChatGPT.} Although we ask ChatGPT to output key semantics that best distinguish between two classes, ChatGPT may generate incorrect semantics in some cases. For instance, in Sample 8 of Table \ref{table:images2}, ChatGPT provides the same semantic for both American Pit Bull Terrier and Abyssinian, which fails to aid CLIP in correcting the original prediction. This issue arises from LLM's capacity limitations to give the key semantics.
\end{enumerate}
It should be noted that many of the issues mentioned above typically occur only in a minority of hard test cases. In most other test cases, our proposed new semantics have been able to assist CLIP in making correct classifications, as evidenced by the performance improvement of CLIP reflected in Table 1 and Figure 4 of the main text. Moreover, some of these issues can be expected to be effectively resolved in the present or future. 
For instance, eliminating the ambiguity in class names can address many of the failure cases we just mentioned.
With the continuous enhancement of CLIP and Large Language Models, the performance of our method will also improve accordingly.

\section{Analysis of Rerank Stage}
In this section, we will furthe discuss the rerank stage.
\subsection{Reasons of Method Effectiveness}
\autoref{table:images3} illustrates examples of image classification results corrected through the rerank step. Observing the samples in the table, we can see that using the classification scores gap based on one-to-one specific semantics for reranking can correct the original predictions. We believe this is primarily due to three reasons: 
\begin{enumerate}
    \item The semantics that distinctly differentiate a class from various other classes may vary. This leads to the one-to-one specific semantics generated by \ATG having a more comprehensive description of the current class, thereby improving classification performance. For example, the most obvious difference between a ``cheetah'' and a ``cougar'' is whether there are spots on the body, while the most obvious difference between a ``cheetah'' and a ``snow leopard'' is the color of the fur. This means that the one-to-one specific semantics generated for ``cheetah'' need to focus on both the pattern and color of the fur. And these diverse one-to-one specific semantics can help us describe the class more comprehensively.
    \item Our reranking method can be considered as an ensemble of several binary classifiers. These classifiers typically satisfy the "good but different" criterion, ensuring the effectiveness of the ensemble algorithm.
    \item Our method effectively utilizes the quantified information of relative advantages between different classes' classification scores. Unlike voting methods that rerank classes based solely on which class receives more votes, our method better captures the model's uncertainty during the process of class prediction. For instance, a small score difference might indicate the model's lack of certainty between two specific class classifications, suggesting that the corresponding classification result may not be reliable. Therefore, it leads to errors in the answers derived from voting methods.
\end{enumerate}
\subsection{The Choice of K}
In the reranking stage, we rerank the top $K$ classes from the initial classification results of the first stage. Here $K$ is a hyperparameter. 
Since only those images whose true class are within the top $K$ of the initial prediction results may potentially be corrected by our method.
Therefore, in general, the larger the value of $K$, the more images that can potentially be corrected, and the better the performance of the method. However, the total number of one-to-one specific test sets involved in the reranking process of the current image is also increasing, leading to greater costs and computational expenses. Therefore, the choice of $K$ requires a trade-off between performance and expenses. In our paper, we set $K=5$, as we find that for many image classification tasks, CLIP can already largely ensure that the true class of the image is among the top 5 preliminary predicted classes. It should be noted that due to the varying matching capabilities of CLIP for different image-text pairs, as well as the limitations of ChatGPT~(as the previous analysis of bad cases of one-to-one specific semantics in \ref{failure_case}), a smaller value of $K$ might yield better results than a larger $K$ in some cases.
\subsection{Complexity of Rerank Stage}
The one-to-one specific texts are created at the class granularity instead of instance granularity and, once constructed for one class, can be saved and reused for future image classifications.
We analyze the computational complexity of the classification process for a single image.
Consider there are $N$ classes, $K$ general texts per class, and $K$ one-to-one specific texts per class pair. The feature dimension of CLIP is $d$. For simplicity, we focus only on the multiplications for computing similarity scores between image and text features. 
For the first stage, we need to perform the inner product between the image's feature vector and the feature vectors of $N \times K$ texts. This requires $N \times K \times d^{2}$ multiplications.
For the second stage, we need to perform the inner product between the image's feature vector and the feature vectors of $C_5^2$ one-to-one texts. This requires $C_5^2 \times K \times d^{2} =10\times K \times d^{2}$ multiplications. 
Thus, the total multiplication is $(N + 10) \times K \times d^{2}$, indicating the complexity of the reranking process is small. 

\section{More Discussion about \CODER}
In this section, we provide further discussion about our \CODER.
\subsection{Comparison with Related Work}
we discuss the differences and innovations of our work compared to some related works.
\vspace{10pt}

\noindent\textbf{Compared with Concept Bottleneck Models.} Some previous works~\cite{cbm1,cbm2} try to train a linear model on the image's concept scores to complete the image classification task. This model is known as the \textbf{C}oncept \textbf{B}ottleneck \textbf{M}odel~(CBM). Its classification process is based on the weighting sum of various concept scores for the images, which gives it good interpretability. The calculation of these concept scores is based on the computation of cross-modal image-text similarities using CLIP. Our work primarily differs from these methods in terms of key ideas and methods.

For key ideas, previous works about CBM mainly leverage the image-text match scores for model interpretability, while our method uses these scores to construct the image's neighbor representation for boosting CLIP’s performance. We reveal that this neighbor representation can compensate for the deficiencies in CLIP’s original features while those works about CBM lack such discussion or opinion. 

For method, those work~\cite{cbm1,cbm2} also use Large Language Models~(LLMs) to get texts with different semantics, but our method differs from them in the implementation detail. Their approach first uses a simple query prompt to get a large candidate text set, then filters it to get a discriminative and diverse subset. However, their filter methods require cumbersome hyperparameter tuning~\cite{cbm2} or image-based training~\cite{cbm2}. Moreover, their methods may fail to direct LLM to produce diverse texts. In contrast, our method directly guides LLM to produce diverse, high-quality features, bypassing the need for cumbersome filtering, by employing diverse query templates. Our method is simple, without the need for images, filtering process, training process, and hyperparameter tuning process, while still generating diverse texts.
\vspace{10pt}

\noindent\textbf{Compared with External Knowledge-based CLIP Inference Methods.} Previous works~\cite{vcd,cupl} have demonstrated the potential of transferring knowledge from external experts like LLMs to enhance CLIP’s performance during the inference stage. However, what knowledge should be transferred from LLMs to better aid models remains an open question with vast research potential. And our \ATG has undertaken new explorations for this question. \ATG introduces three new text types, including Analogous Class-based Texts, Synonym-based Texts, and One-to-One Specific Texts. Our \ATG achieves more beneficial knowledge transfer from LLMs to CLIP, with experiment results showing consistent improvements over previous method~\cite{vcd}.

\noindent\textbf{Comparison with CLIP training-free few-shot image classification methods.} Previous works~\cite{tip-adapter,sus-x} have also attempted to enhance CLIP's few-shot performance through a training-free approach. Among these, TIP-Adapter~\cite{tip-adapter} and TIP-X~\cite{sus-x} both try to utilize the similarity between test images and few-shot images to correct the original CLIP's zero-shot classification results based on text classifiers.
The difference is that TIP-Adapter directly uses the image features extracted by CLIP's image encoder to calculate the similarity between images. While TIP-X computes the similarity using each image's concept score vectors between the image and the semantics generated by CuPL~\cite{cupl}. 
Our main difference from these methods is that we are the first to interpret image-text matching scores as images’ neighbor representations. 
From this perspective, we can conclude that to construct better neighbor representations, we need a diverse and high-quality text set, as suggested by the dense sampling condition of the nearest neighbor algorithm.
We demonstrate through experiments that increasing the diversity and quantity of texts can improve the quality of images' neighbor representations, thereby enhancing the performance of previous clip training-free few-shot image classification methods, such as TIP-Adapter.
Previous works did not include our perspective of understanding the image-text matching scores as images' neighbor representations, nor did they discuss how to construct high-quality image neighbor representations in the CLIP feature space, while we provide the answer --- a high-quality and diverse text set is required to meet the dense sampling conditions.

\subsection{Limitations}
Our method has several limitations:

The first limitation is the cost of generating texts using LLMs. When there are a large number of classes, we will also need a relatively large number of texts generated by \ATG. This might require more frequent calls to the ChatGPT API for generation, thereby incurring higher API call costs and longer generation times. 

The second limitation is the issue of \CODER's dimension. The dimension of \CODER is equal to the number of class-related texts generated by \ATG, thus it is proportional to the number of classes. When there are few classes, the small number of texts may make it difficult to meet the dense sampling conditions for the construction of images' \CODER, thus affecting the quality of \CODER.
On the other hand, when there are many classes, the feature dimension of \CODER might become excessively high, leading to increased computational resource costs when calculating images' similarities based on \CODER. This problem can be resolved through dimension reduction. However, the dimension reduction process also introduces additional time and computational costs.

Lastly, due to the limitations of CLIP or LLM capabilities, the rerank stage may not be able to correctly modify CLIP's prediction results for images in some cases.

Regarding the first and third limitations mentioned above, the solution we employed in our experiment is that we only use rerank stage for images with a high probability of error in the initial classification results.
We determine whether the initial predicted class of the image is incorrect by calculating the difference between the largest and the second largest logits values in the initial classification logits predictions for the image. 
The smaller the difference, the greater the uncertainty of the model's prediction, indicating a higher likelihood of an incorrect prediction.
And for those images with a larger difference in values, we consider CLIP's classification result to be highly likely correct.
We then compare this difference with a preset threshold value, and only the test images with differences smaller than this threshold are subject to re-ranking. 
 For the generation of one-to-one specific texts, we employ a dynamically constructed method: If the one-to-one specific texts needed at the moment have been previously generated, we reuse the previous results. Conversely, we utilize \ATG to generate the one-to-one specific texts and then save them.

We will further optimize these limitations in subsequent work.

\onecolumn
\begin{longtable}[H]{| m{1cm}<{\centering}|m{13cm}<{\centering} |}
\hline
Sample ID & Instance \\
\hline
1 & \includegraphics[width=\linewidth]{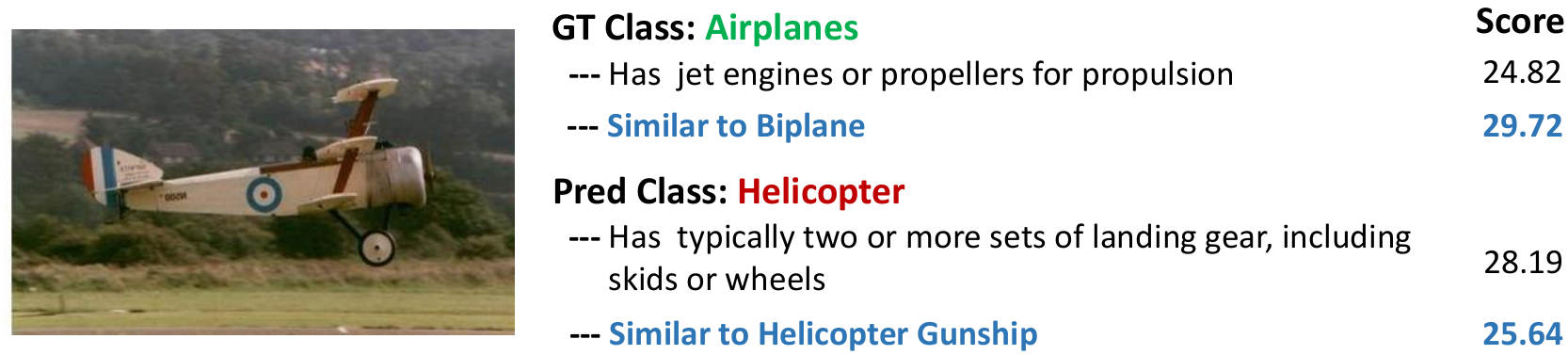} \\
\label{1.1}\\
\hline
2 & \includegraphics[width=\linewidth]{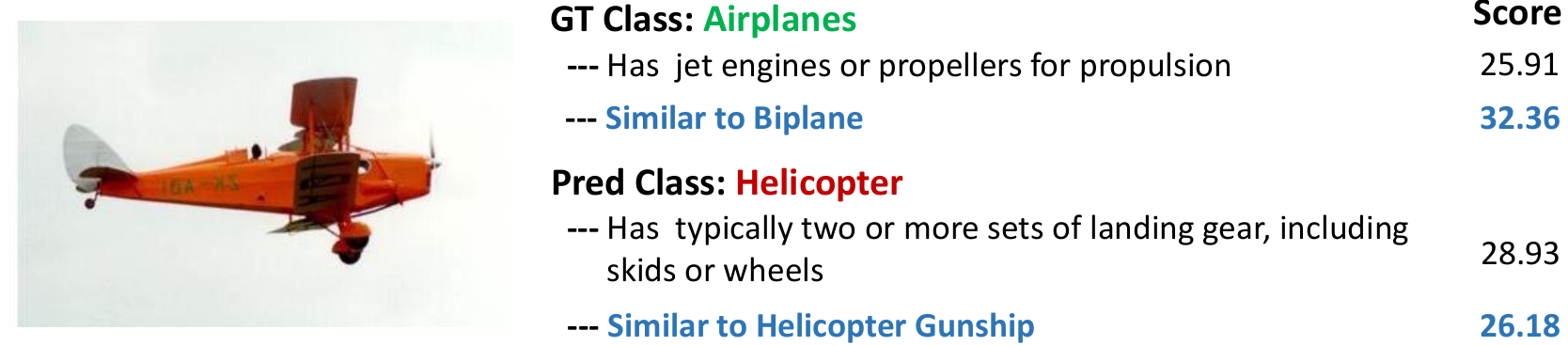} \\
\hline
3 & \includegraphics[width=\linewidth]{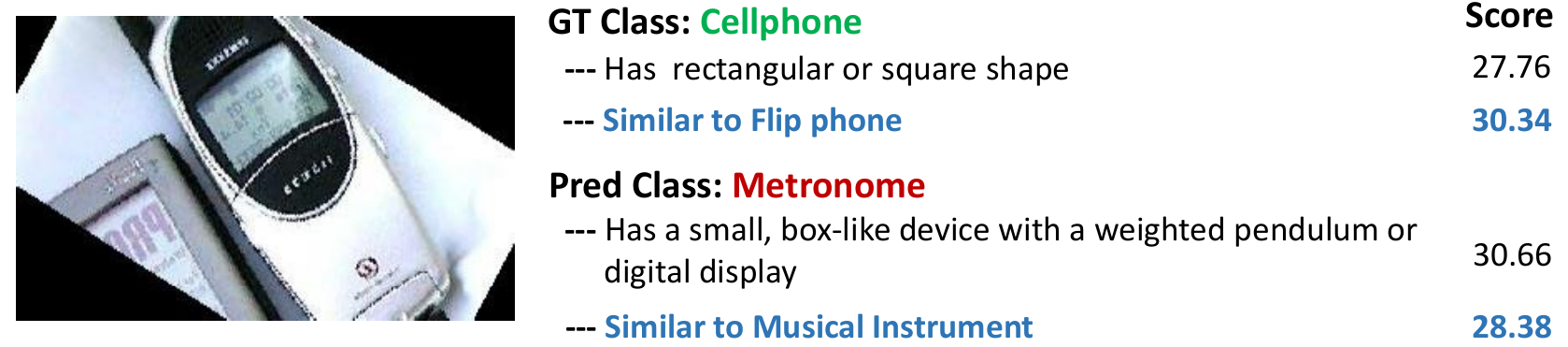} \\
\hline
4 & \includegraphics[width=\linewidth]{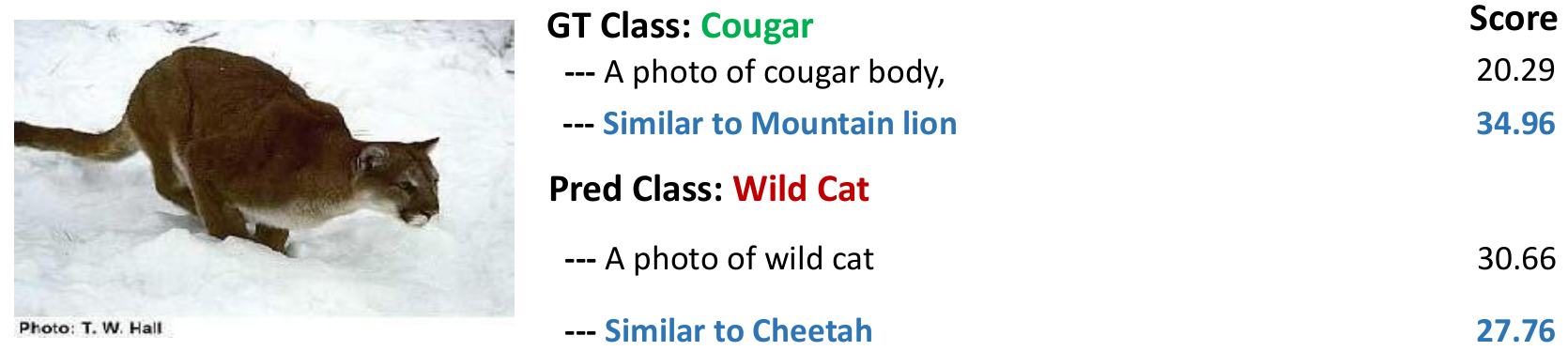} \\ 
\hline
5 & \includegraphics[width=\linewidth]{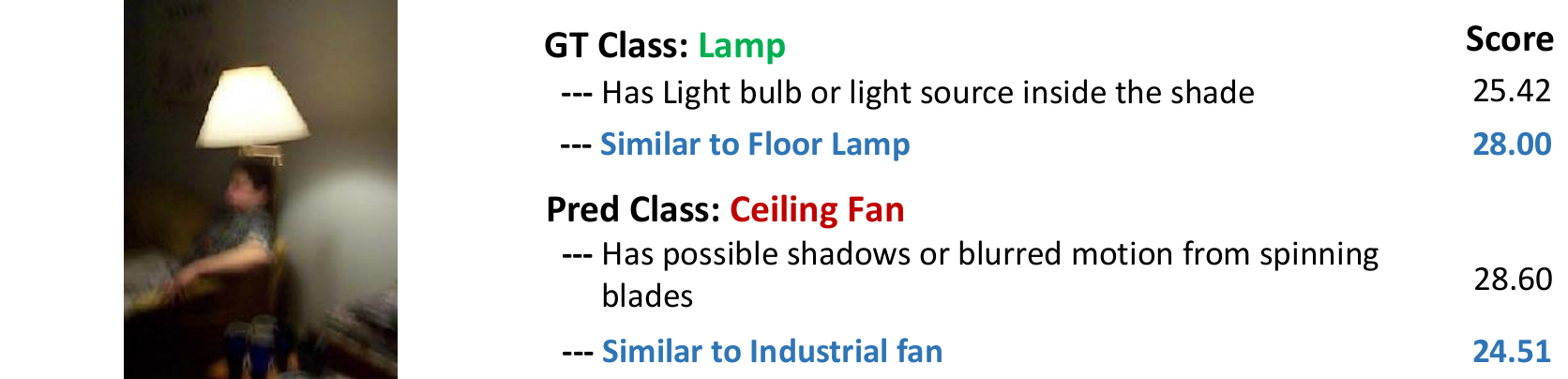} \\
\hline
6 & \includegraphics[width=\linewidth]{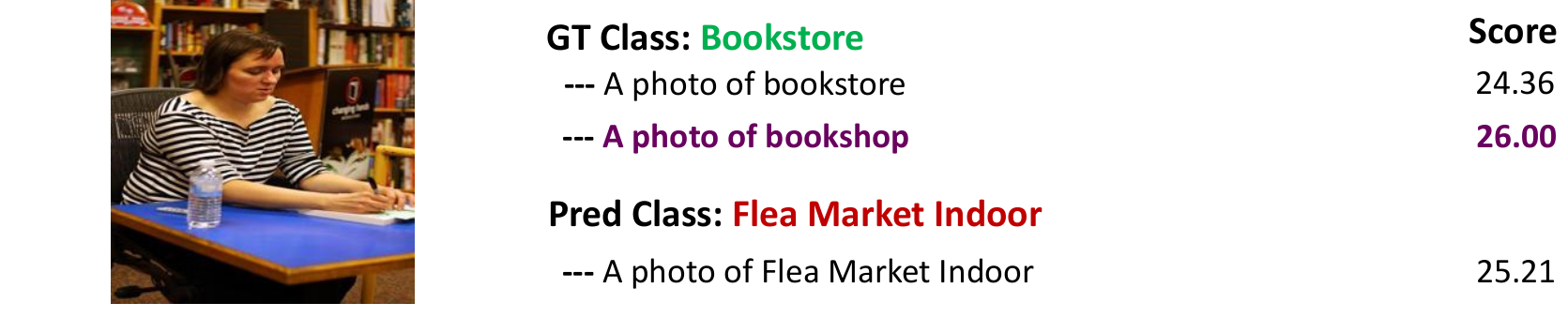} \\
\hline
7 & \includegraphics[width=\linewidth]{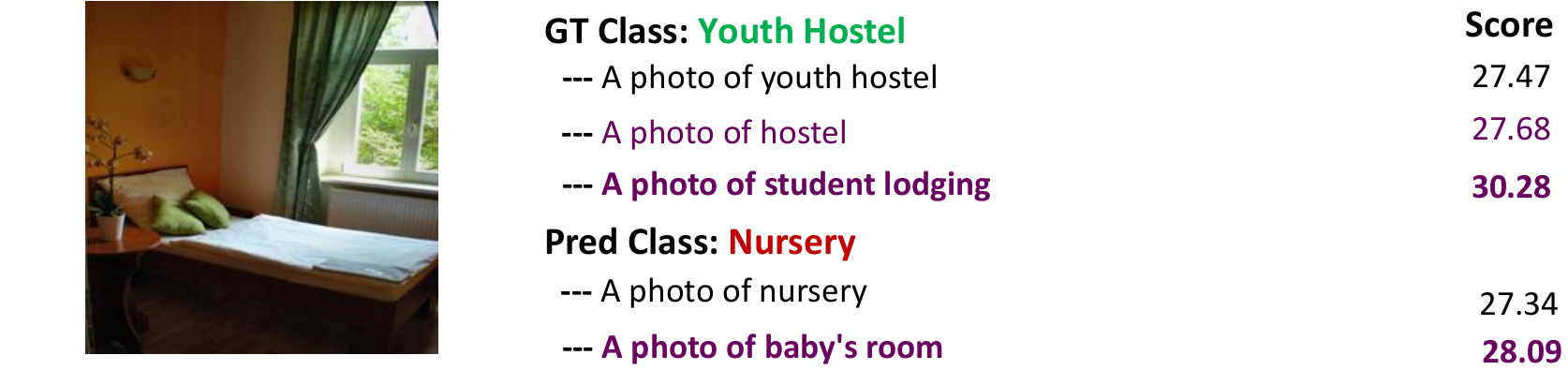} \\
\hline
8 & \includegraphics[width=\linewidth]{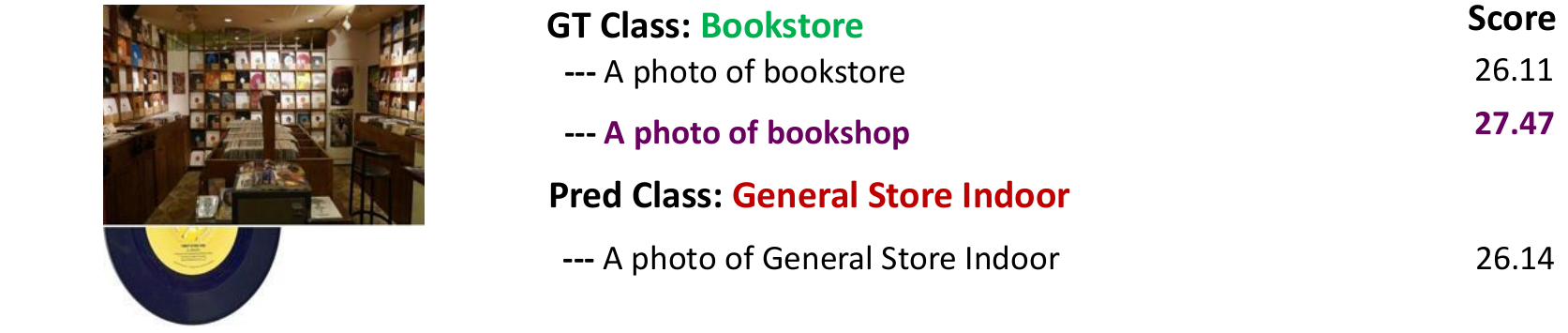} \\
\hline
9 & \includegraphics[width=\linewidth]{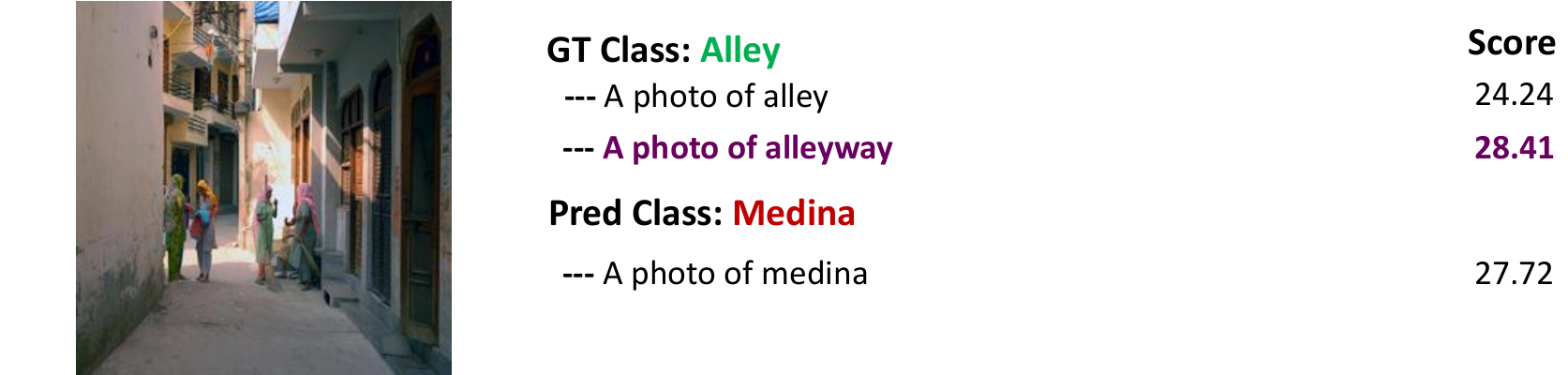} \\
\hline
10 & \includegraphics[width=\linewidth]{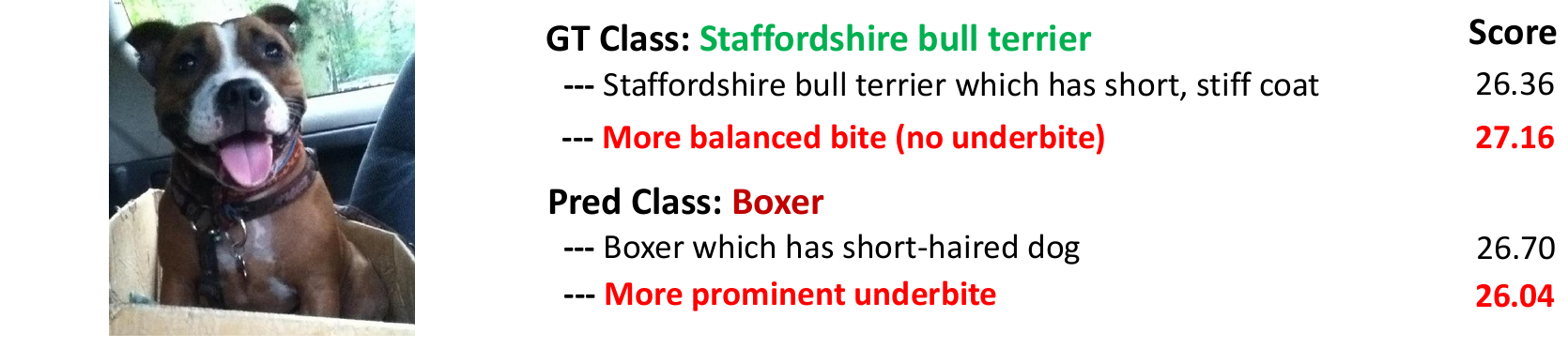} \\
\hline
11 & \includegraphics[width=\linewidth]{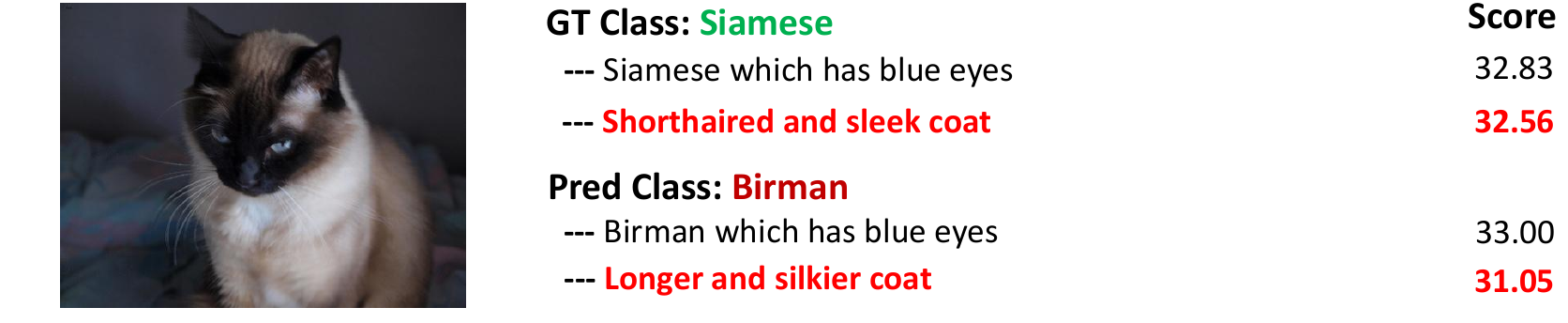} \\
\hline
&\\
12 & \includegraphics[width=\linewidth]{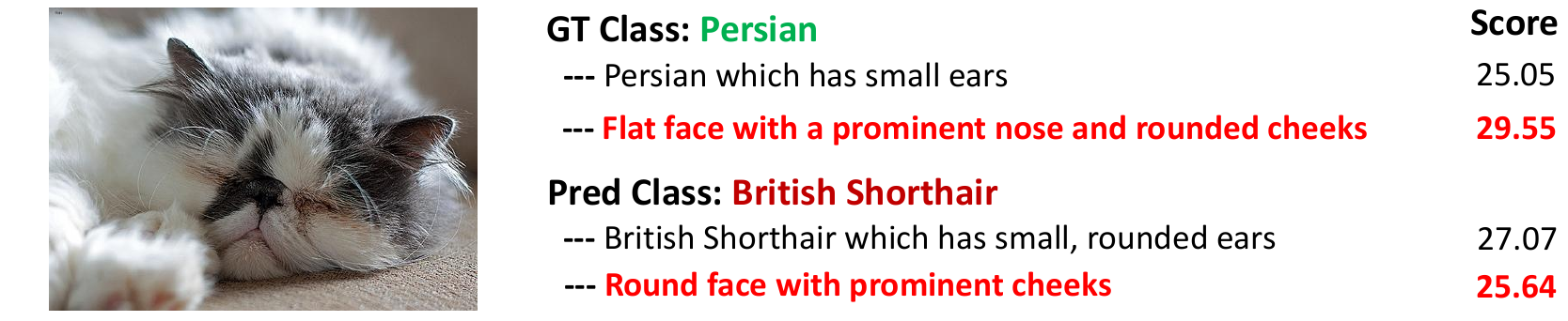} \\
\hline
\caption{\textbf{Examples of using new types of semantics generated by the Auto Text Generator to correct erroneous original predictions in CLIP's zero-shot image classification.} The figure demonstrates the different images' matching scores with semantics for each test image's original CLIP-predicted class versus its true class. Here, the black text represents the previous semantics~(including CLIP's original class name-based semantics and attribute-based semantics from VCD~\citep{vcd}). \textcolor{blue}{Blue text} signifies our proposed analogous class-based semantics. \textcolor{purple}{Purple text} signifies synonym-based semantics, and \textcolor{red}{red text} signifies one-to-one specific semantics. The multitude of examples in the figure prove that our newly proposed semantic types can enhance CLIP's classification capabilities.}
\label{table:images}\\
\end{longtable}
\twocolumn

\onecolumn
\begin{longtable}[H]{| m{1cm}<{\centering}|m{13cm}<{\centering} |}
\hline
Sample ID & Instance \\
\hline
1 & \includegraphics[width=\linewidth]{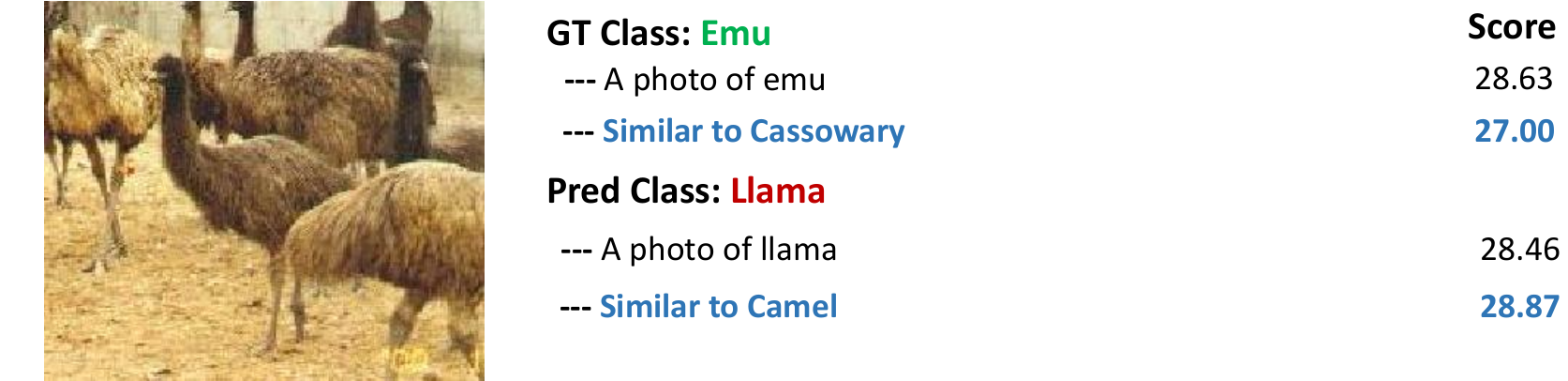} \\
\hline
2 & \includegraphics[width=\linewidth]{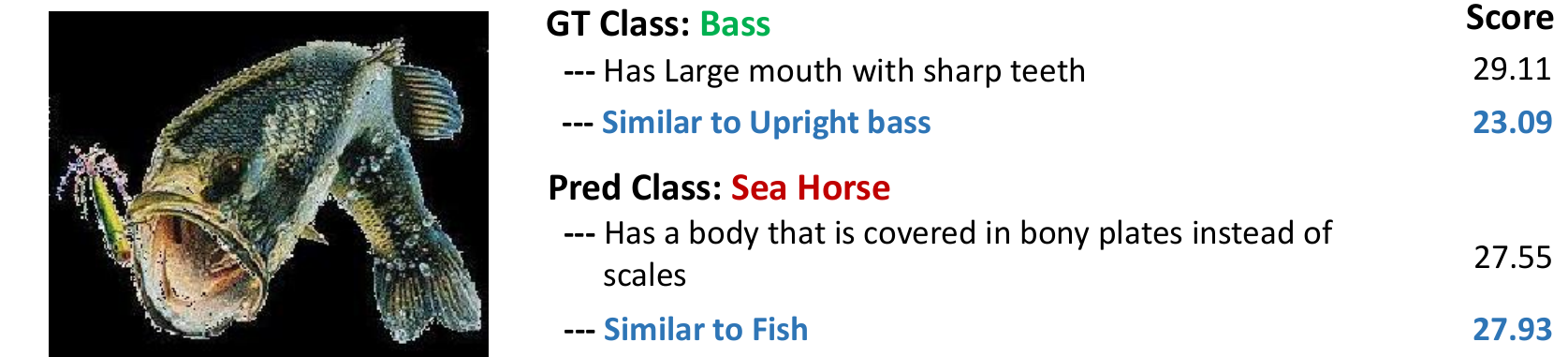} \\
\hline
3 & \includegraphics[width=\linewidth]{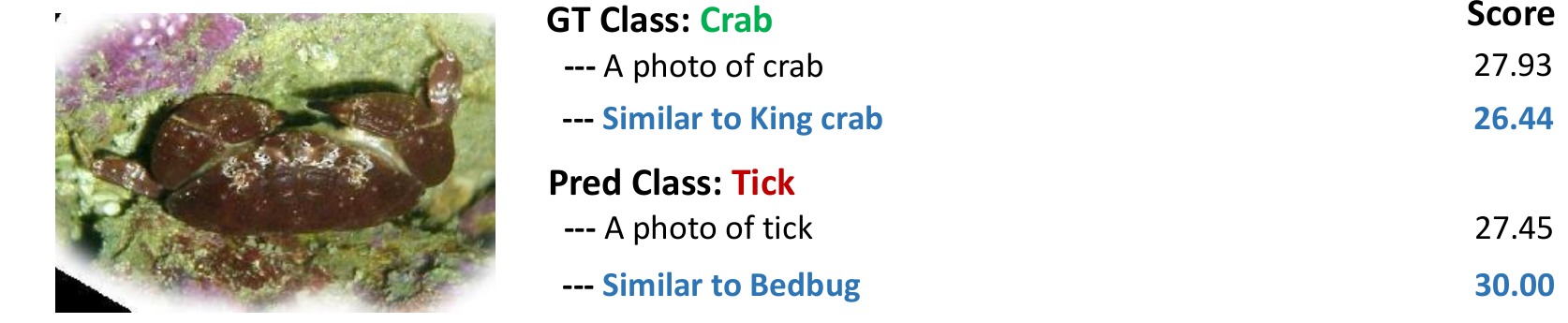} \\
\hline
4 & \includegraphics[width=\linewidth]{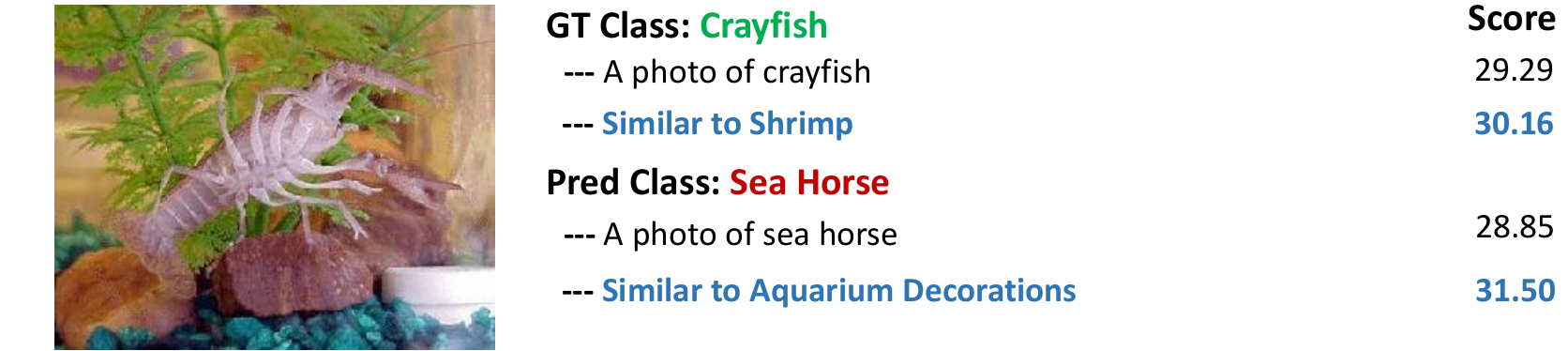} \\
\hline
5 & \includegraphics[width=\linewidth]{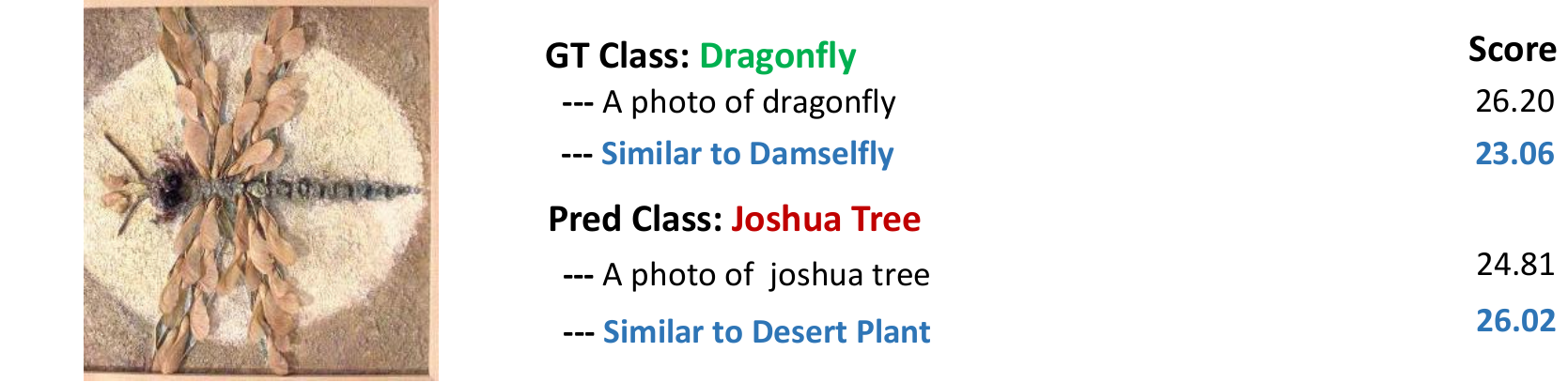} \\
\hline
6 & \includegraphics[width=\linewidth]{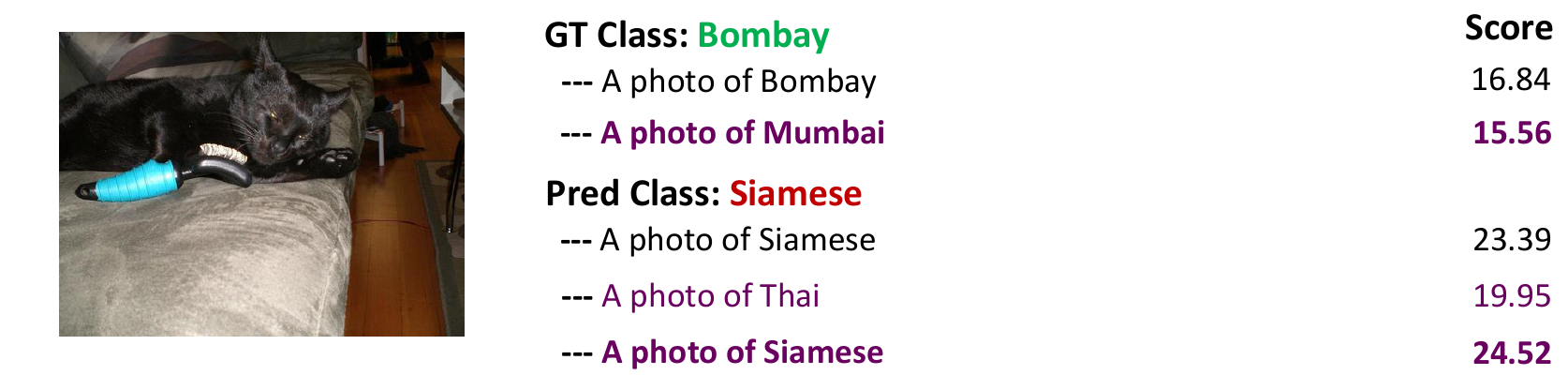} \\
\hline
7 & \includegraphics[width=\linewidth]{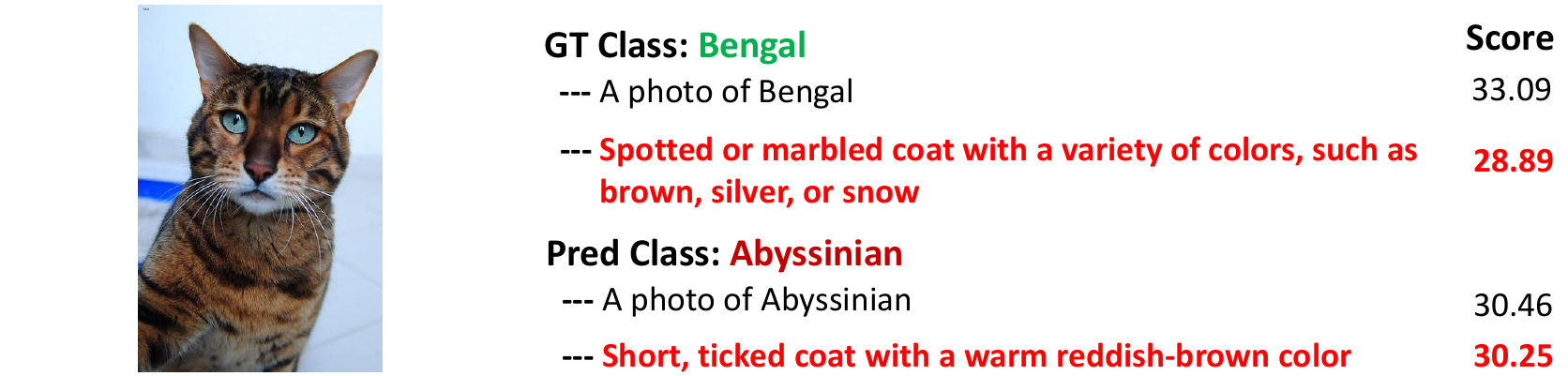} \\
\hline
8 & \includegraphics[width=\linewidth]{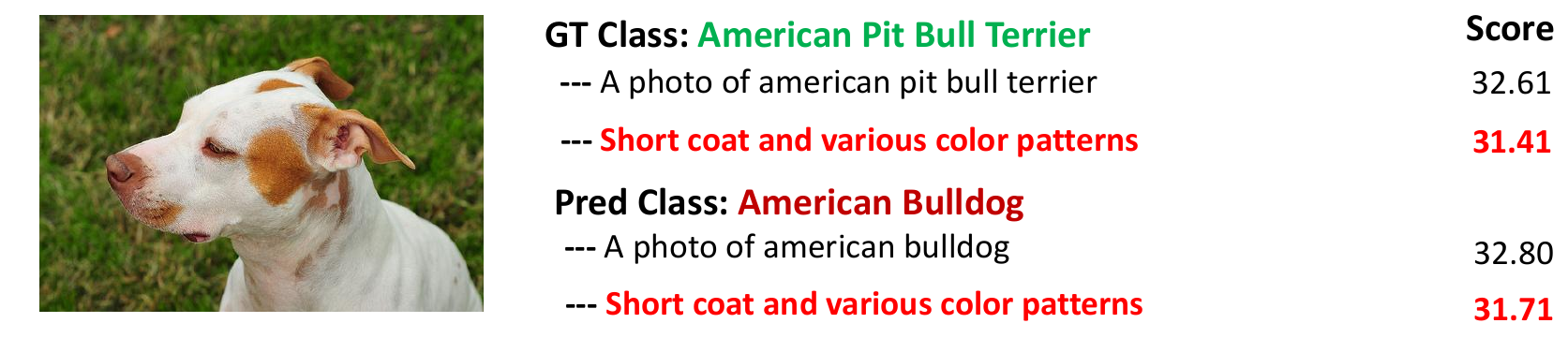} \\
\hline
\caption{\textbf{Bad case examples in zero-shot image classification using new semantics generated by the Auto Text Generator.} Here, black text represents the previous semantics~(including CLIP's original class name-based semantics and attribute-based semantics from VCD~\citep{vcd}). \textcolor{blue}{Blue text} signifies our proposed analogous class-based semantics. \textcolor{purple}{Purple text} signifies synonym-based semantics, and \textcolor{red}{red text} signifies one-to-one specific semantics.}
\label{table:images2}\\
\end{longtable}
\twocolumn

\onecolumn
\begin{longtable}[H]{| m{1cm}<{\centering}|m{13cm}<{\centering} |}
\hline
Sample ID & Instance \\
\hline
1 & \includegraphics[width=\linewidth]{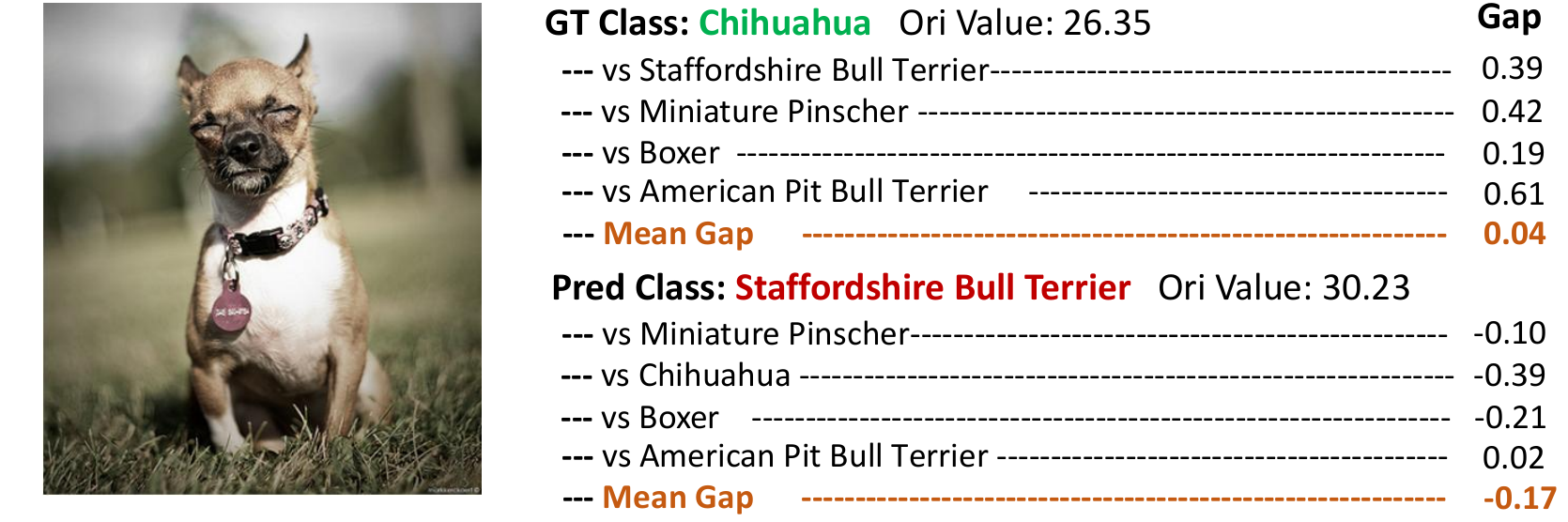} \\
\hline
&\\ 
\hline
2 & \includegraphics[width=\linewidth]{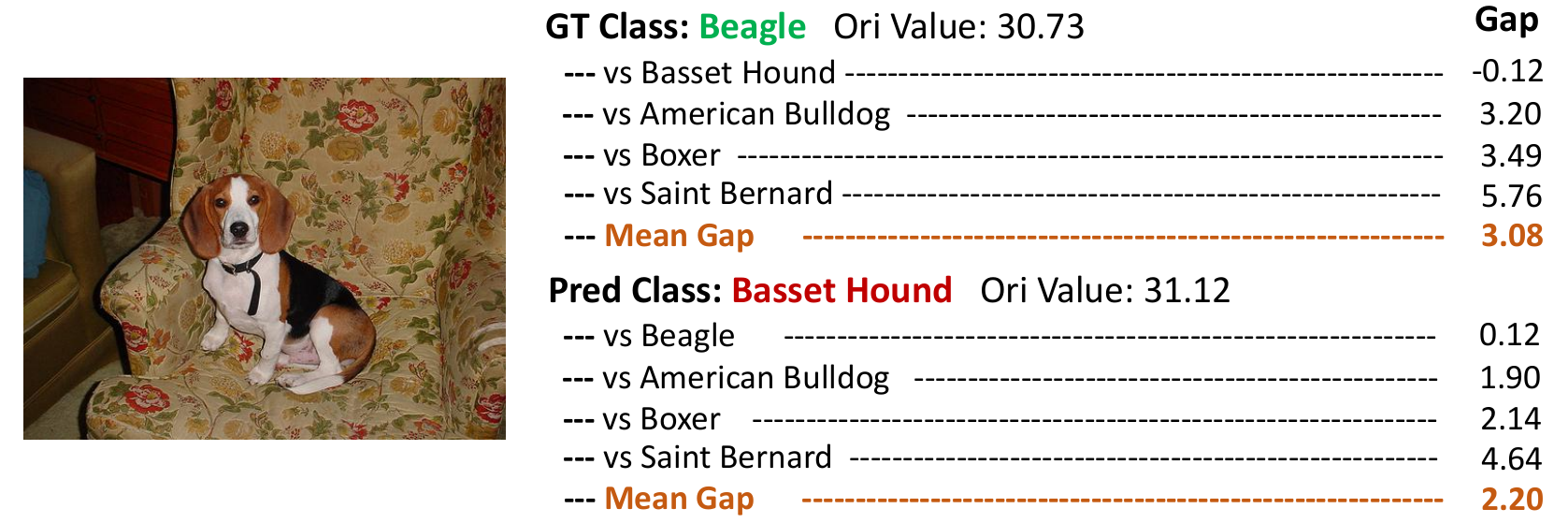} \\
\hline
&\\ 
\hline
3 & \includegraphics[width=\linewidth]{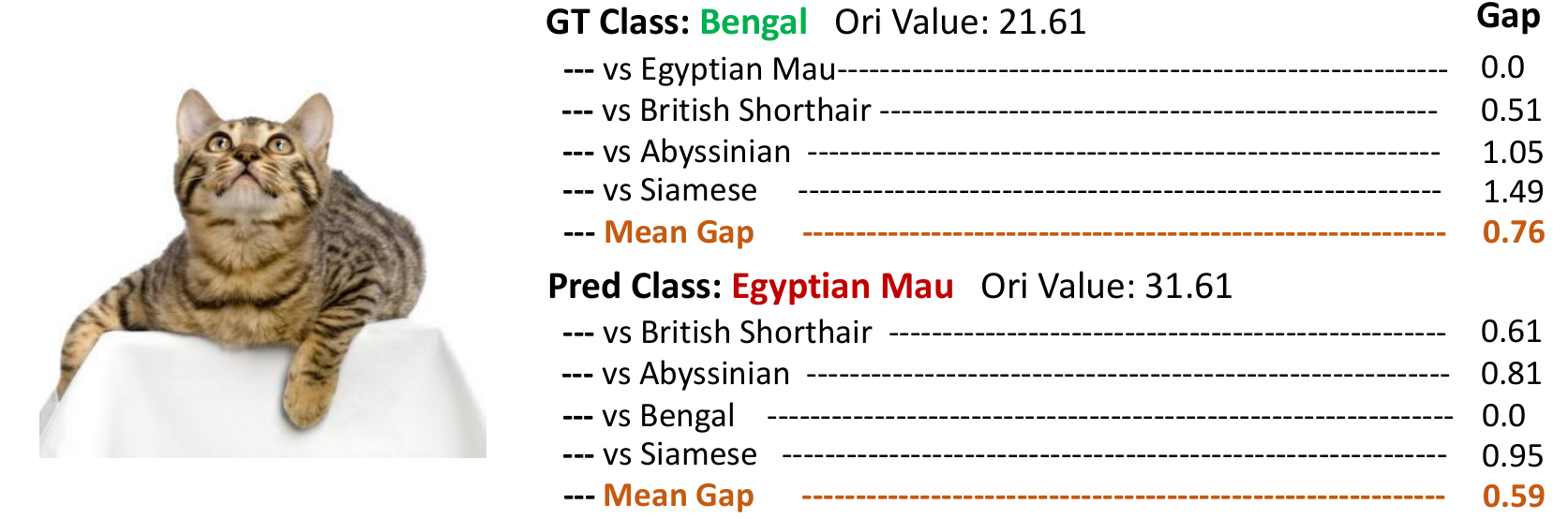} \\
\hline
&\\ 
\hline
4 & \includegraphics[width=\linewidth]{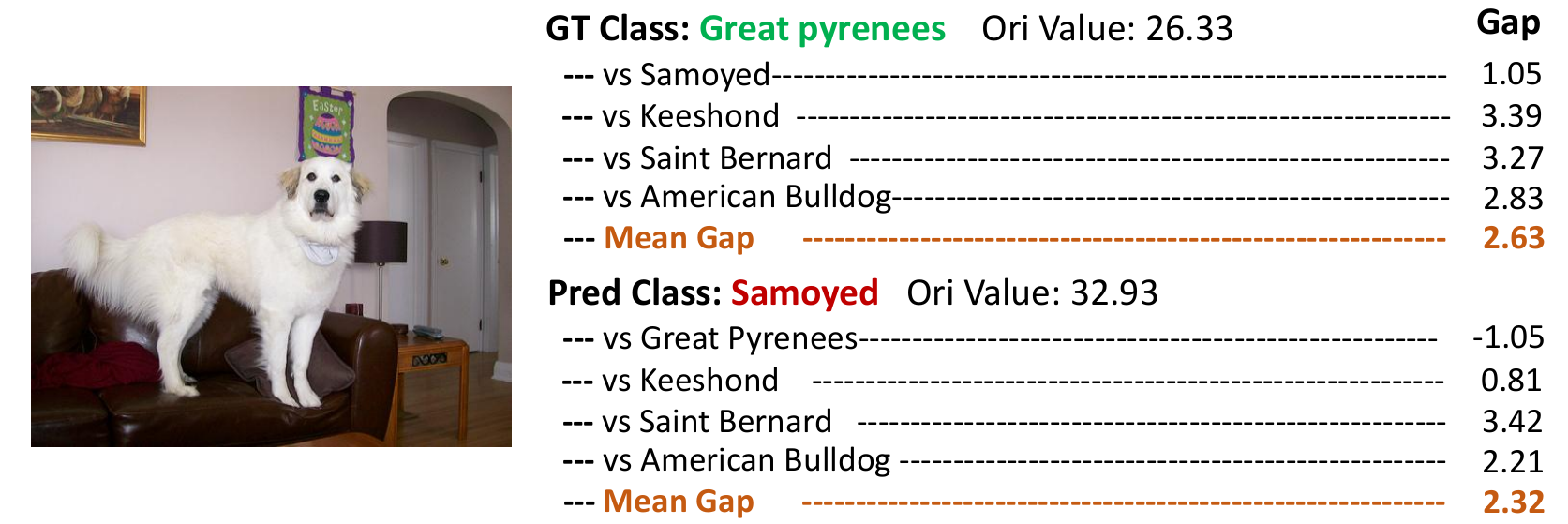} \\
\hline
&\\
\hline
5 & \includegraphics[width=\linewidth]{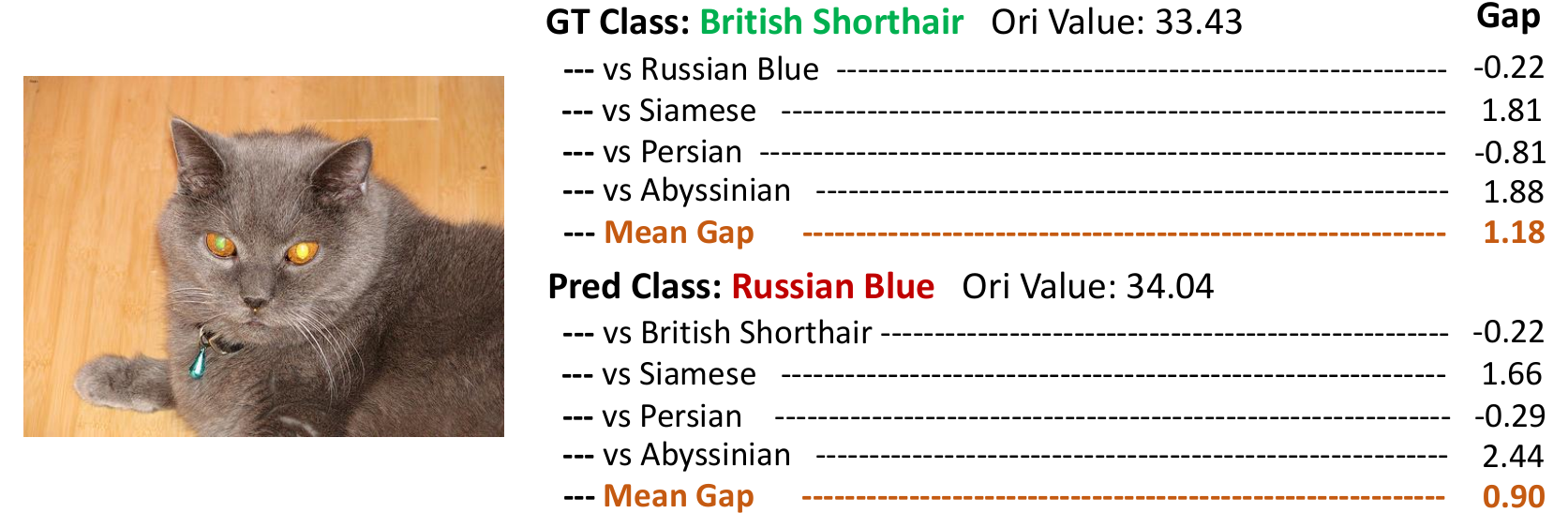} \\
\hline
&\\ 
\hline
6 & \includegraphics[width=\linewidth]{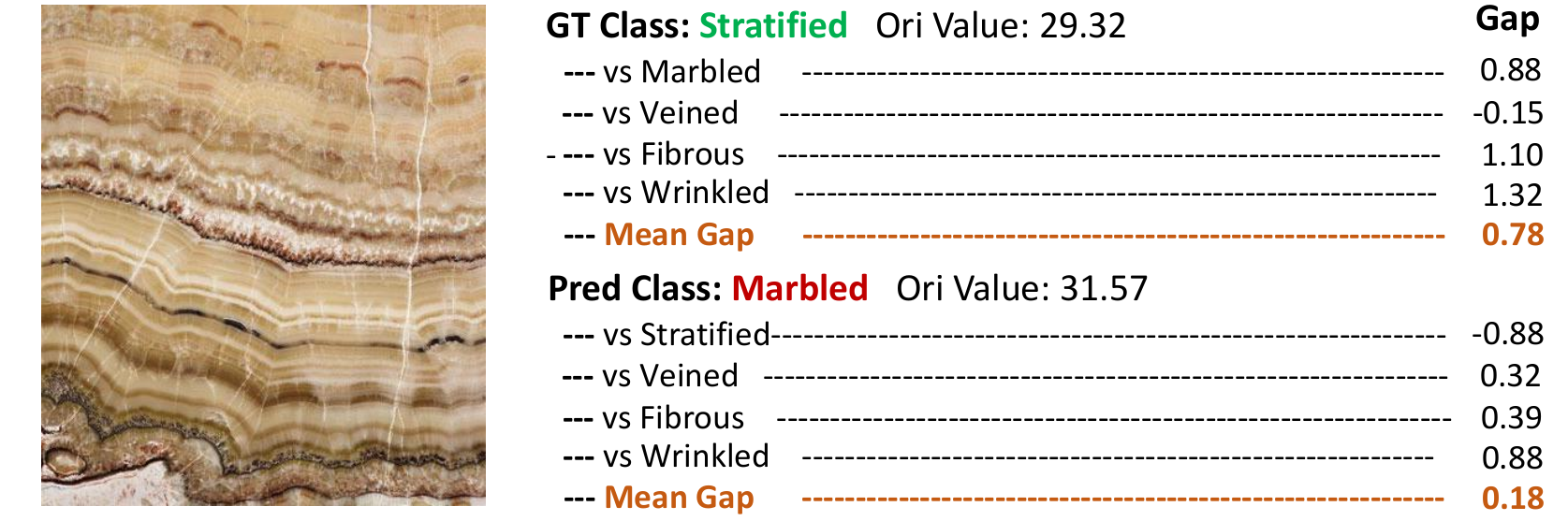} \\
\hline
&\\ 
\hline
7 & \includegraphics[width=\linewidth]{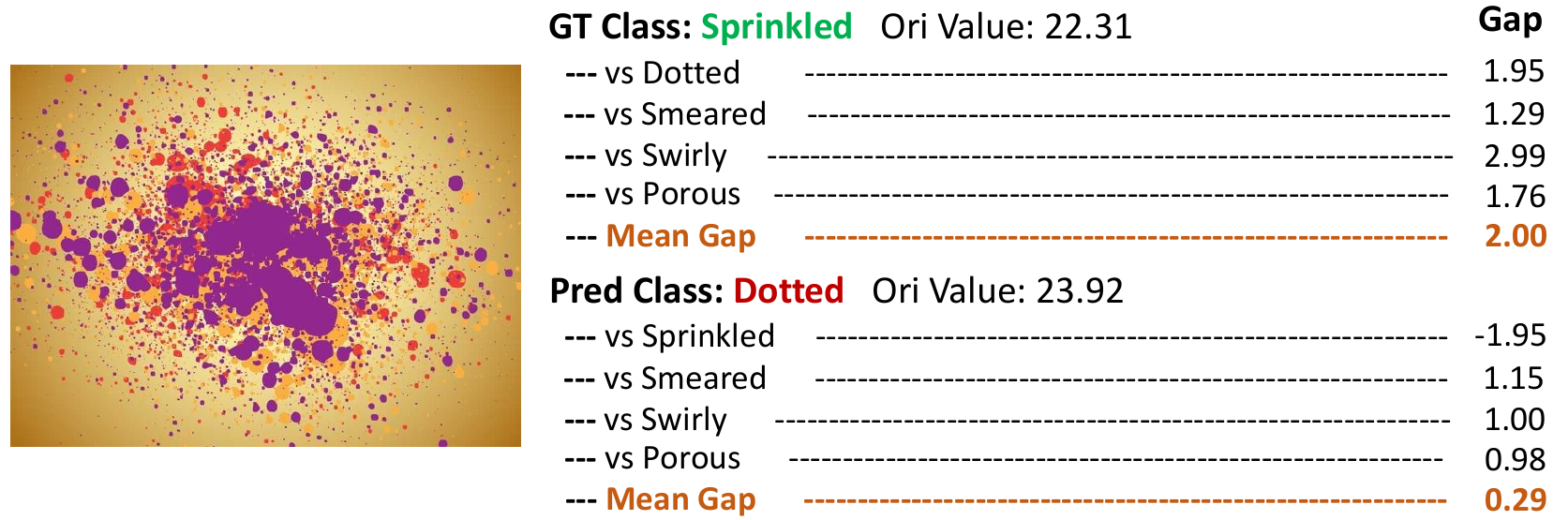} \\
\hline
&\\ 
\hline
8 & \includegraphics[width=\linewidth]{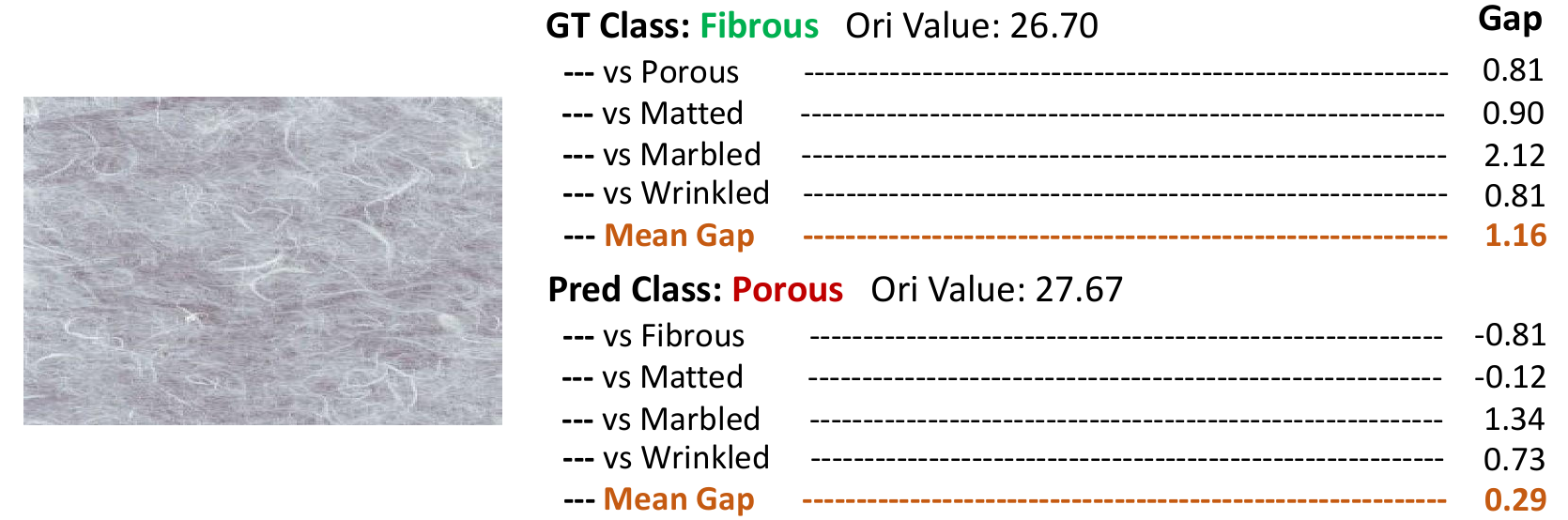} \\
\hline
&\\
\hline
9 & \includegraphics[width=\linewidth]{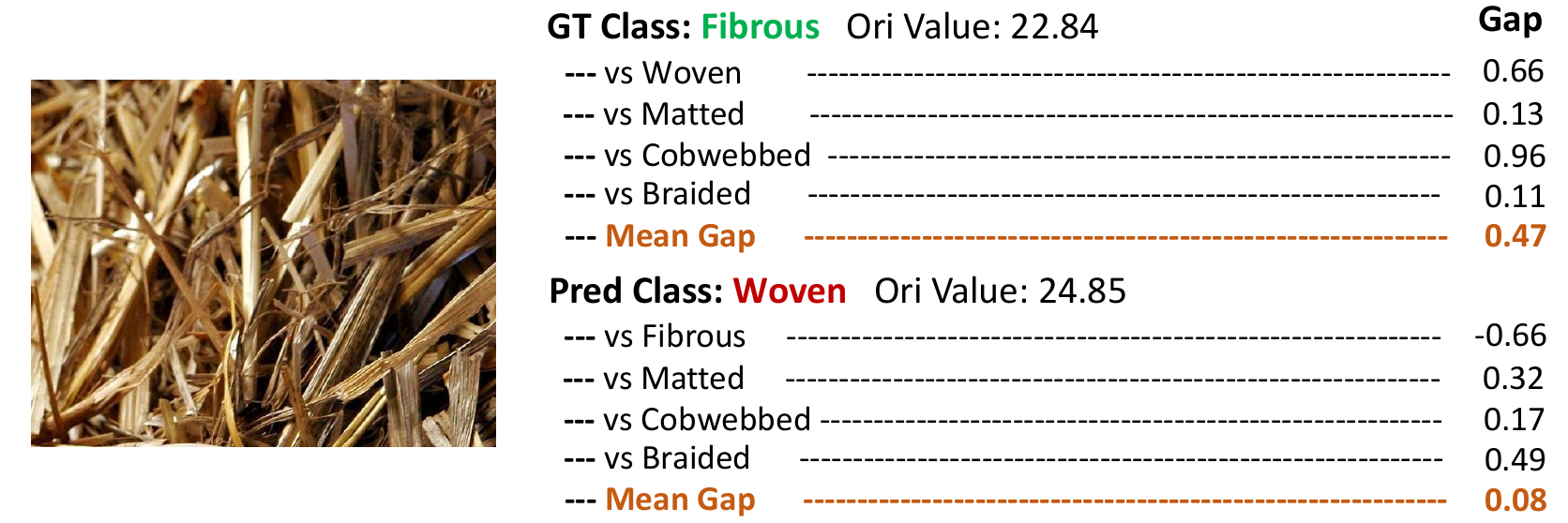} \\
\hline
&\\ 
\hline
10 & \includegraphics[width=\linewidth]{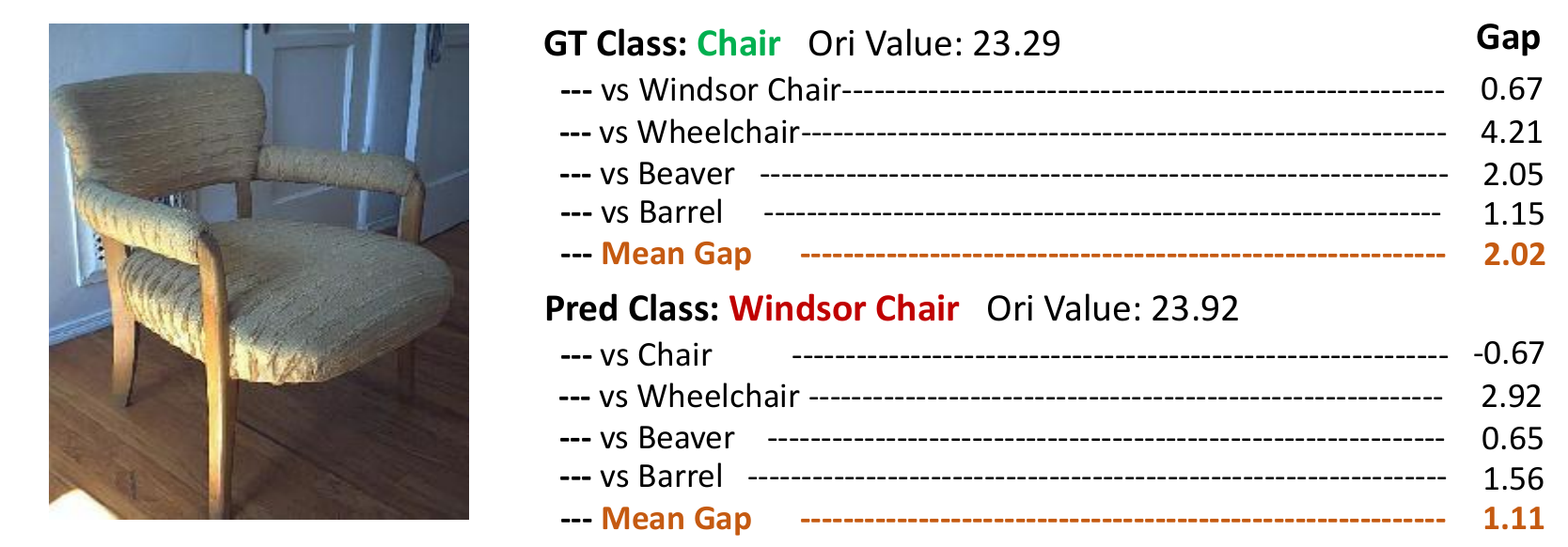} \\
\hline
&\\ 
\hline
11 & \includegraphics[width=\linewidth]{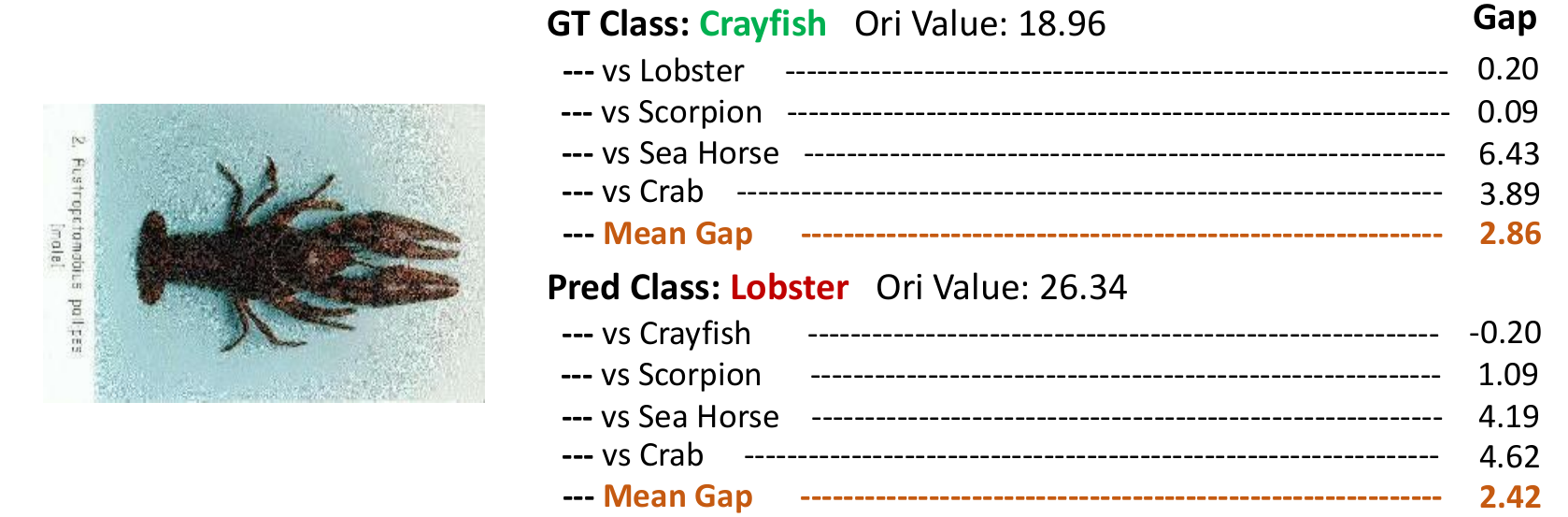} \\
\hline
&\\ 
\hline
12 & \includegraphics[width=\linewidth]{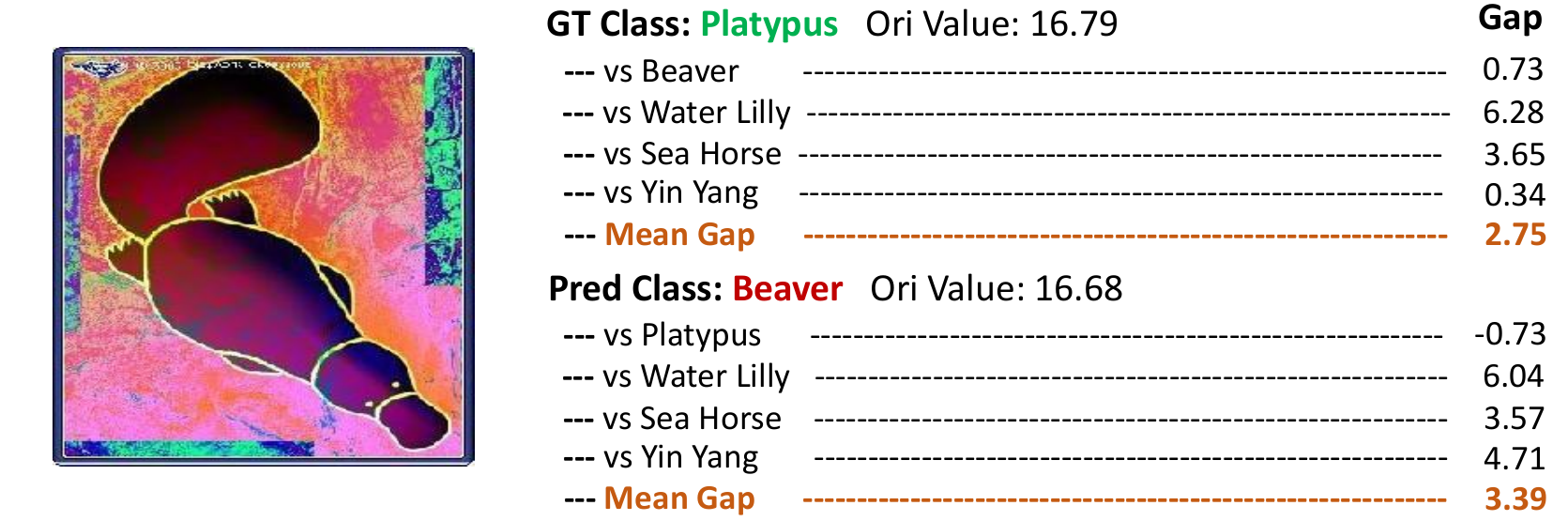} \\
\hline
&\\
\hline
13 & \includegraphics[width=\linewidth]{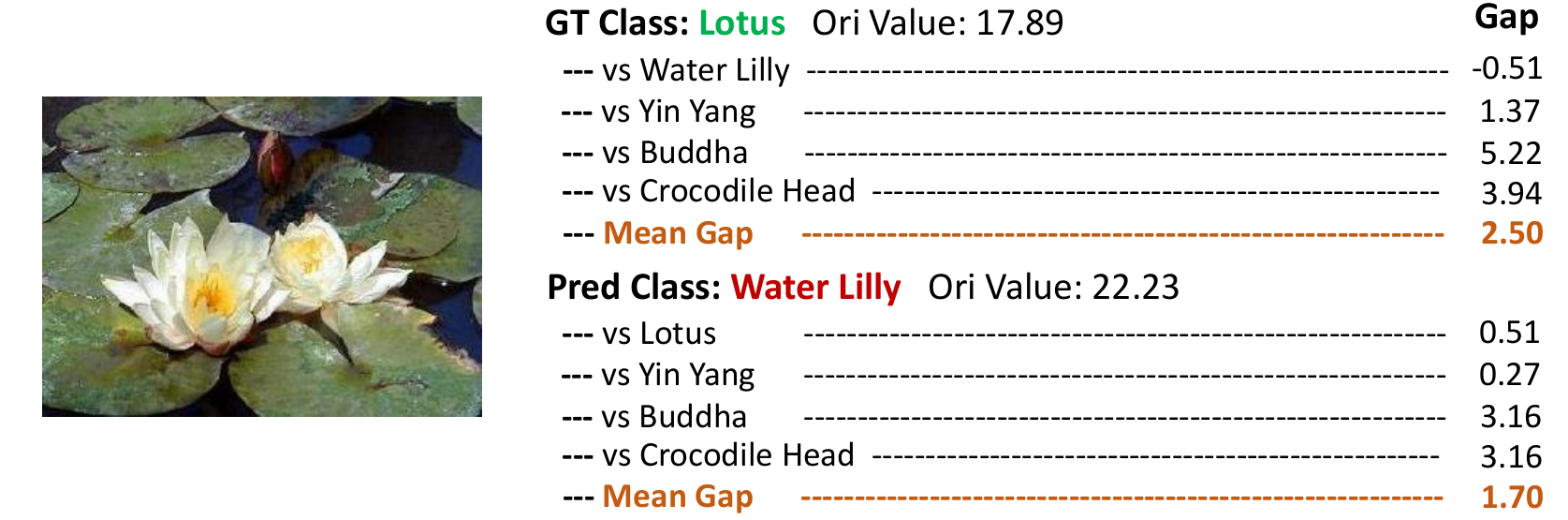} \\
\hline

\caption{\textbf{Examples of correcting original classification results through reranking based on One-to-One Specific \CODER.} By introducing one-to-one semantics to accentuate features with the maximum differences between classes, and leveraging the rich information contained in the relative strengths of predictive scores across classes, we successfully rectify the original wrong predicted results.}
\label{table:images3}\\
\end{longtable}
\twocolumn

\end{document}